\documentclass[sigconf]{acmart}
\AtBeginDocument{%
  }

\setcopyright{acmlicensed}
\copyrightyear{2018}
\acmYear{2018}
\acmDOI{XXXXXXX.XXXXXXX}
\acmConference[Conference acronym 'XX]{Make sure to enter the correct
  conference title from your rights confirmation email}{June 03--05,
  2018}{Woodstock, NY}
\acmISBN{978-1-4503-XXXX-X/2018/06}

\usepackage{algorithm}
\usepackage{algorithmic}
\usepackage{multicol}
\usepackage{multirow}
\usepackage{booktabs}
\usepackage{rotating}
\usepackage{color}
\usepackage{subcaption}
\usepackage{amsthm}                
\usepackage{amsfonts}
\usepackage{pifont}

\usepackage{xcolor}
\usepackage{newfloat}
\usepackage{dirtytalk}
\usepackage{listings}

\usepackage{pgfplots}
\pgfplotsset{compat=newest}
\usepackage{tikz}
\newtheorem{theorem}{Theorem}

\begin{document}

\title{Lifting Manifolds to Mitigate Pseudo-Alignment in LLM4TS}

\author{Liangwei Nathan Zheng}
\affiliation{%
  \institution{The University of Adelaide}
  \city{Adelaide}
  \country{Australia}
}
\email{liangwei.zheng@adelaide.edu.au}

\author{Wenhao Liang}
\affiliation{%
  \institution{The University of Adelaide}
  \city{Adelaide}
  \country{Australia}
}
\email{wenhao.liang@adelaide.edu.au}

\author{Wei Emma Zhang}
\affiliation{%
  \institution{The University of Adelaide}
  \city{Adelaide}
  \country{Australia}
}
\email{wei.e.zhang@adelaide.edu.au}

\author{Miao Xu}
\affiliation{%
  \institution{The University of Queensland}
  \city{Adelaide}
  \country{Australia}
}
\email{miao.xu@uq.edu.au}

\author{Olaf Maennel}
\affiliation{%
  \institution{The University of Adelaide}
  \city{Adelaide}
  \country{Australia}
}
\email{olaf.maennel@adelaide.edu.au}

\author{Weitong Chen}
\authornote{Corresponding Author}
\affiliation{%
  \institution{The University of Adelaide}
  \city{Adelaide}
  \country{Australia}
}
\email{weitong.chen@adelaide.edu.au}



\renewcommand{\shortauthors}{Anonymous et al.}

\begin{abstract}
Pseudo-Alignment is a pervasive challenge in many large language models for time series (LLM4TS) models, often causing them to underperform compared to linear models or randomly initialised backbones. However, there is limited discussion in the community for the reasons that pseudo-alignment occurs. In this work, we conduct a thorough investigation into the root causes of pseudo-alignment in LLM4TS and build a connection of pseudo-alignment to the cone effect in LLM. We demonstrate that pseudo-alignment arises from the interplay of cone effect within pretrained LLM components and the intrinsically low-dimensional manifold of time-series data. In addition, we also introduce \textit{\textbf{TimeSUP}}, a novel technique designed to mitigate this issue and improve forecast performance in existing LLM4TS approaches. TimeSUP addresses this by increasing the time series manifold to more closely match the intrinsic dimension of language embeddings, allowing the model to distinguish temporal signals clearly while still capturing shared structures across modalities. As a result, representations for time and language tokens remain distinct yet exhibit high cosine similarity, signifying that the model preserves each modality’s unique features while learning their commonalities in a unified embedding space. Empirically, TimeSUP consistently outperforms state-of-the-art LLM4TS methods and other lightweight baselines on long-term forecasting performance. Furthermore, it can be seamlessly integrated into four existing LLM4TS pipelines and delivers significant improvements in forecasting performance. Code is released in : \url{https://github.com/IcurasLW/LLM4TS_Mitigate_Pseudo_Alignment.git}
\end{abstract}



\received{20 February 2007}
\received[revised]{12 March 2009}
\received[accepted]{5 June 2009}

\maketitle

\section{Introduction}
\begin{figure*}[ht]

    \centering
    \begin{subfigure}[b]{\linewidth}
        \centering
        \includegraphics[width=\linewidth, trim=0 2em 0 0]{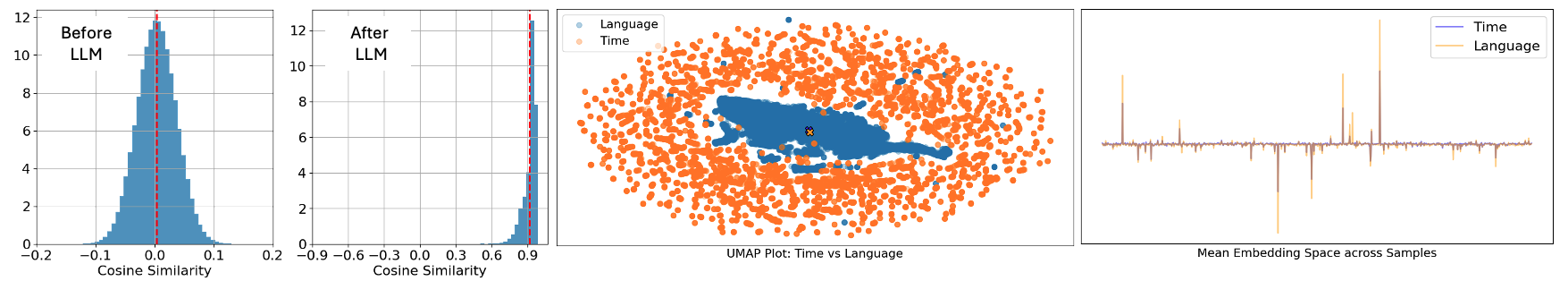}
        \caption{Cone Effect and UMAP Representation without Manifold Enhancer}
        \label{fig:no-mixer}
    \end{subfigure}
    \hfill
    \begin{subfigure}[b]{\linewidth}
        \centering
        \includegraphics[width=\linewidth, trim=0 2em 0 0]{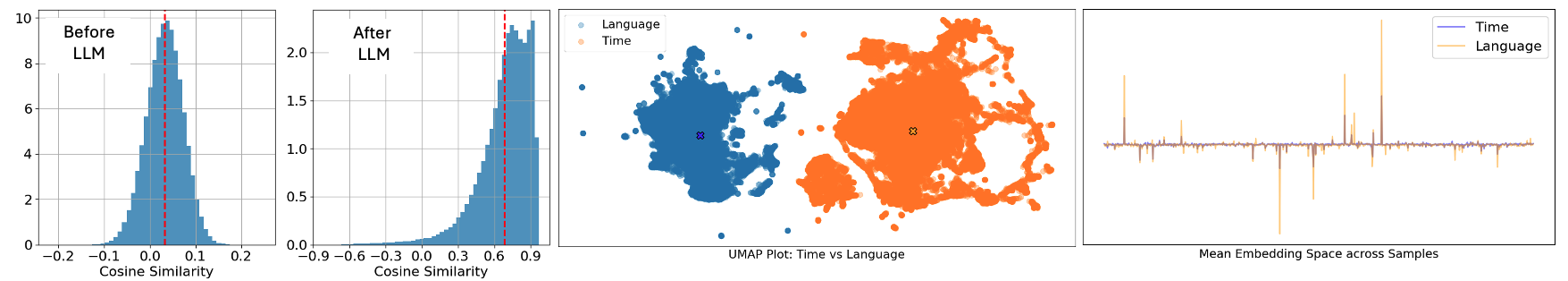}
        \caption{Cone Effect and UMAP Representation with Manifold Enhancer}
        \label{fig:with-mixer}
    \end{subfigure}
     \captionsetup{skip=2pt}
    \caption{Comparison between the representation and similarity with and without TimeSUP: (1) Cosine similarity is determined by paring time series and language tokens. (2) Pseudo-Alignment is still preserved if time series token is fed in LLM. (3) Pseudo-Alignment phenomenon is mitigated by TimeSUP.}
    \label{fig:compare-mixer}
\end{figure*}

Large Language Model has been shown to have significant impacts on World-Wide-Web application such as real-time online decision making by Time Series analysis, ranging from online medical diagnosis \cite{zheng2024irregularity} to weather \cite{mudelsee2010climate}, traffic flow \cite{zheng2020traffic}, and energy-load forecasting \cite{sharadga2020time}.  Recently, large language models (e.g., GPT-2 and LLaMa) have been repurposed for forecasting via prompting, LoRA fine-tuning, and LayerNorm adaptation, giving rise to various LLM4TS (Large Language Models for Time Series) techniques. However, because these models are exclusively pre-trained in text, their suitability for time series tasks has come under criticism \cite{zheng2024revisited,tan2024language} and they argue that a replacement of LLM to linear backbone or random initialized weight LLM can win a lot and even draw to LLM backbone. This challenges the premise that pretrained language knowledge directly benefits time series forecasting. A recent work \cite{zheng2024revisited} attributed this ineffectiveness of LLMs to an empirical phenomenon termed \say{Pseudo-Alignment}. This refers to a situation where \textit{model representations of time-series and language appear superficially aligned at the level of first-order statistics (e.g., mean, see Figure \ref{fig:no-mixer}, most right), yet their full distributions remain distinct, revealing a failure of true semantic alignment and a distortion of modality-specific features.} The persistence of pseudo-alignment has been shown to consistently degrade model performance, sometimes performing even worse than linear or randomly initialized backbones \cite{zheng2024revisited,tan2024language}. Consequently, there has been limited systematic investigation into the mechanistic origins of pseudo-alignment in LLM4TS models. Closing this gap is essential not only for fair evaluation of current approaches but also for guiding future designs.It motivates two central questions that remain largely unanswered: \textbf{(1) What may be the root causes for pseudo-alignment?} and \textbf{(2) Which architectural modifications or training strategies can effectively activate the rich knowledge of LLMs for time series forecasting?} Addressing these questions can provide a principled foundation for future research and help distinguish when, how, and why language modality can be genuinely useful in time series domains.

\begin{figure}[h]
  \centering
  \includegraphics[width=\linewidth]{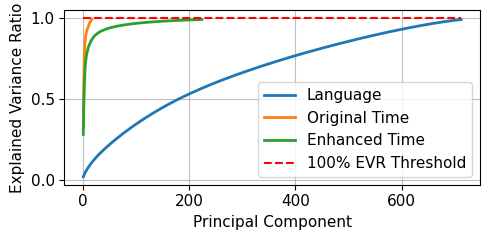}
  \caption{PCA Explained Variance Ratio plot for Language, Original, and Enhanced Time series}
  \label{fig:PCA_EVR}
\end{figure}

Although \cite{zheng2024revisited} empirically suggests that the performance of time series task come from the internal structure of time series modality instead of language due to the existence of pseudo-alignment, our systematic investigation further reveals the underlying causes of pseudo‑alignment additional to the internal structure views and implement comprehensive exploratory analysis to find the following observations: (1) pseudo‑alignment is a synthetic effect of the LLM cone effect \cite{gao2019representation} coupled with the inherently low‑dimensional manifold of time‑series data, (2) the architectural components primarily responsible triggering Pseudo-Alignment are the pretrained LayerNorm and Multi‑Head Attention layers within the LLM, and (3) because of the cone effect, LLMs tend to \say{take easy route} by enforcing \say{perfect} alignment rather than recognizing the true low‑dimensional structure of time‑series. With these observations, we respect the rich language knowledge of LLM can contribute to time series analysis \cite{zheng2024revisited,jin2023time} and theoretically show that low dimensionality of time series can cause pseudo-alignment. High dimensionality manifold of time series representation can prevent LLM from pseudo-alignment during the propagation of cone effect. Ideally, we expect the input time series to match the manifold dimension as language so as to activate the rich knowledge of LLM, and an ideal alignment exhibit preservation of distinct modality discrepancy while maintaining high cosine similarity in their shared subspace, thereby retaining modality‑specific features and aligning shared information. The high discrepancy indicates the model can clearly tell the difference of modality distribution that provide unique modality feature and structure, which is highly correlated to modality gap phenomenon as discussed in \cite{liang2022mind}. The high cosine similarity at token-level indicates that the time series and language modality lives in a consistent and highly aligned embedding space \cite{gao2019representation}.

In addition, we also introduce a simple, yet can broadly adapted module for many LLM4TS approaches to effectively address pseudo-alignment and boost the performance in long-term time series forecasting task. Based on observations (3), we proposed TimeSUP, a Time Manifold Enhancer, that elevates the manifold dimensionality of time‑series data (originally 21 dimensions) toward that of GPT‑2’s language token manifold (712 dimensions), yielding strong and consistent performance improvements in our experiment. To demonstrate the effects of this dimensional enhancement, we conducted a probe experiment with PCA (see Figure \ref{fig:PCA_EVR}). Results show that the enhanced representation raises the intrinsic manifold dimension of time series from 21 to 224 in order to preserve a 0.99 explained variance ratio, thereby significantly increasing the data manifold. This dimensional expansion allows the time series to occupy a higher-dimensional manifold that is more compatible with the pretrained LLM’s representational geometry. Further visualization in Figure \ref{fig:compare-mixer} (second column) reveals that the pseudo-alignment issue is notably mitigated, while token similarity remains at relatively high values. This indicates that TimeSUP not only preserves local relational structures but also avoids the distortion of modality-specific features, offering extensible insights into modality alignment. Our findings suggest that dimension enhancement is a critical mechanism to unlock the latent capabilities of LLMs in non-linguistic domains. By mitigating pseudo-alignment, the proposed Time Manifold Enhancer provides a generalizable solution that can be integrated into a wide range of LLM4TS pipelines. This line of exploration opens new opportunities for bridging modality gaps and further harnessing the power of LLMs for time series understanding and beyond. Our main contributions can be summarized as follows:

\begin{itemize}
    \item We are the first work to reveal pseudo-alignment problem from the perspective of data manifold dimensionality, revealing new insights of LLM4TS model. We also implement comprehensive experiment to show the impact of low dimensionality of time series.
    \item We proposed \textit{\textbf{TimeSUP}}, an simple, yet effective way of reprogramming large language model for time series, to effectively address Pseudo-Alignment problem by raising the exact dimensionality of time series data. Similar learning paradigm can be extendable to other form of reprogramming large language model for time series.
    \item TimeSUP consistently outperforms the state-of-art LLM4TS baseline on various long-term forecasting dataset. Moreover, it is easily adaptable to other LLM4TS approach, achieving instant performance improvement over original baseline implementation.
\end{itemize}

\section{Preliminary}

\subsection{Time Series Lies in a Lower-Dimensional Data Manifold} \label{sec:time_in_lower}

According to the manifold hypothesis \cite{cayton2005algorithms}, real-world data often reside on low-dimensional manifolds embedded within high-dimensional spaces. This suggests that complex observations can be described using only a few intrinsic dimensions. To examine this, we apply PCA to both time-series tokens $T_i \in \mathbb{R}^d $ and GPT-2 language tokens $L_i \in \mathbb{R}^d$, where $d=768$. Our results show that time-series tokens (generated with a patch size of 16 and stride of 8) can be well-represented using just 21 principal components, whereas language tokens retain 712 out of 768 components. This indicates that time-series modality lie on a much lower-dimensional manifold than language modality \cite{liu2012robust}.

This observation helps explain why pretrained language models can still perform well on time-series tasks under the model reprogramming assumption: if the source model is trained on a more complex domain (language), it can be adapted to a simpler target domain (time series). In essence, successful cross-modal reprogramming relies on the pretrained model being expressive enough to capture the target distribution \cite{englert2022adversarial,elsayed2018adversarial,yang2021voice2series}. However, the simplicity of the time-series manifold may also limit the generalization ability of LLMs in time-series applications, as the representation of time series may be easier to manipulated than complex representation.

\subsection{Cone Effect in Large Language Models}

The Cone Effect refers to a phenomenon where token embeddings in LLMs collapse into a narrow cone, reducing their representational diversity \cite{gao2019representation,liang2022mind,bis2021too}. Initially observed in language token embeddings, this effect causes tokens to become increasingly similar cosine similarity values rise close to 1 after each LLM layer, whereas the tokens were initially well-dispersed in the feature space. This is largely due to architectural components like LayerNorm and Multihead Attention \cite{shukor2024implicit}. The cone effect can reduce the expressiveness of LLMs, as all tokens tend to appear positively correlated, even if they are semantically distinct. In the setting of multimodal learning, when inputs from different modalities are fed into the model, multimodal data may form separate cones with a gap in between due to data heterogeneity, which can positively impact downstream performance \cite{liang2022mind}.

In the context of LLMs for time series (LLM4TS), we also clearly observe this cone effect. As shown in Figure \ref{fig:no-mixer}, the mean cosine similarity between language and time tokens is near zero and spans mostly in between -0.1 to 0.1, but significantly increases beyond 0.9 after LLM and most of token similarity are all positive. The UMAP visualization and mean representation plot also illustrate a strong Pseudo-Alignment, where time-series representations collapse toward those of language in a centroid, rather than achieving meaningful cross-modal alignment. Ideally, we hypothesis that a well-aligned model should maintain distinct modality-specific features while aligning shared semantics as they are partially shared information across modality instead of perfectly mapping relationship, resulting in separate yet high token similarity in the shared embedding space \cite{liang2022mind,gao2019representation}.

\section{Understand Pseudo-Alignment}
In this section, we investigate the underlying cause of Pseudo-Alignment in current LLM4TS methods. We consider GPT-2 with LayerNorm fine-tuning as the representative architecture for our empirical analysis, as it is the most commonly adopted backbone across many LLM4TS approaches \cite{zhou2023one,pan2024textbf,hu2025context,hu2025context}. Our analysis focuses on the long-term time series forecasting task, utilizing the official implementation of OFA \cite{zhou2023one} as the baseline method for visualization, where time series tokens are directly fed into the LLM with only the LayerNorm layers fine-tuned. Through the analysis above, we understand that time-series data lives in a much lower-dimensional manifold space and exhibits a simpler structure than language. Thanks to the pretrained LayerNorm and multi-head attention layers, LLM can manipulate and redistribute the feature map of perceptual tokens (here, time-series) into the textual feature domain. However, the time-series manifold is too simple that the model \say{takes easy route}, and it learns to manipulate features toward perfect, 100\% alignment rather than deeply distinguishing modality-specific and shared information. This manipulation can be considered as an implicit multimodal alignment from pretrained attention and layernorm weight \cite{shukor2024implicit}. Consequently, Pseudo-Alignment emerges as a synergistic effect between the cone effect in LLMs \cite{gao2019representation} and the inherently low-dimensional manifold of time series. This issue is especially detrimental to time series data, as its simple structure makes it particularly vulnerable to such distortion manipulation. 

To support our claim above, we show that the Pseudo-Alignment is caused by the low-dimensional data manifold under the condition of cone effect occurring naturally in LLM. Let

\begin{equation}  \label{eq:ts_data}
    \epsilon_{ts} = [v_1, v_2, \cdots, v_m], 
    \quad v_i \sim \mathcal{N}(0, \sigma_{ts}^2)
\end{equation}

\begin{equation}  \label{eq:txt_data}
    \epsilon_{l} = [v_1, v_2, \cdots, v_n], 
    \quad v_j \sim \mathcal{N}(0, \sigma_{l}^2)
\end{equation}

be the stochastic deviations induced by finite sampling from their respective modality distributions. Here $m$ and $n$ denote the intrinsic manifold dimensionalities of the time-series and language modalities. We assume that the time-series $x_{ts}$ and language $x_{l}$ are independently and identically distributed (i.i.d.) w.r.t their modality distributions and can be represented as the sum of the modality mean and sampling perturbations:

\begin{equation} \label{eq:ts_data}
    x_{ts} = \mu_{ts} + \epsilon_{ts}
\end{equation}

\begin{equation} \label{eq:txt_data}
    x_{l} = \mu_{l} + \epsilon_{l}
\end{equation}

We quantify the cone effect by cosine similarity between time series and language modality in Figure \ref{fig:compare-mixer}. Thus, we can obtain the following theorem:

\begin{theorem} \label{theorem_1:low_manifold_casues_pseudo_alignment}
    When the manifold dimensionality $m \to 0$ and $n \to 0$, The cosine similarity tends to converge to the similarity between the mean of time series and language distribution only without the instance-specific feature involved, leading to pseudo-alignment. Formally, we can represent this pseudo-alignment process as follow:

    \begin{equation} \label{eq:expectation_cosine_cone_effect}
        E\left[cos( x_{ts}, x_l)\right] = E \left[ \frac{\langle x_{ts}, x_l \rangle}{\|x_{ts}\| \|x_l\|} \right] = \frac{\mu_{ts} \mu_{l}}{\sqrt{\|\mu_{ts}\|+m\sigma_{ts}} \sqrt{\|\mu_{l}\|+n\sigma_{l}}}
    \end{equation}

\end{theorem}

In practice, as the cone effect occurs naturally in LLM, which the token-level similarity raises significantly after each layer of LLM. As $m \ll n$ and $m\sigma_{ts}$ can be ignored, if the cosine similarity is significantly increased due to cone effect, then Eq. \ref{eq:expectation_cosine_cone_effect} is encouraged to converge to high similarity between $u_{ts}$ and the entire distribution of language, leading to pseudo-alignment.

\textbf{Proof of Theorem 1}. The numerator can be decomposed to $E[\langle x_{ts}, x_l \rangle] = E[(\mu_{ts} + \epsilon_{ts})(\mu_{l} + \epsilon_{l})] = E[ \mu_{ts}\mu_l ] +  E[\mu_{l}\epsilon_{ts}] +  E[\mu_{ts}\epsilon_{l}] +  E[\epsilon_{l}\epsilon_{ts}] = \mu_{ts} \mu_l$ as $\epsilon$ is zero-mean for both modality. For denominator, we know that 

\begin{align}
    E[\|x_{ts}\|^2] &= E[\| \mu_{ts} + \epsilon_{ts} \|^2] \\
    &= E[\|\mu_{ts}\|^2 + 2\langle \mu_{ts},  \epsilon_{ts} \rangle + \|\epsilon_{ts}\|^2] \\
    &= \|\mu_{ts}\| + E[\|\epsilon_{ts}\|^2] = \|\mu_{ts}\| + E[\sum^m_{i=0} \epsilon_{ts, i}^2] \\
    &= \|\mu_{ts}\| + m E[\epsilon_{ts, i}^2] = \|\mu_{ts}\| + m \frac{1}{m} \sum^m_{i=0} \epsilon_{ts, i}^2 \label{eq:law_of_large_num_applied} \\
    &= \|\mu_{ts}\| + m \sigma_{ts}^2 \label{eq:time_expectation}
\end{align}

Law of large number is applied at Eq. \ref{eq:law_of_large_num_applied} as $m \to \infty$, therefore, $\frac{1}{m} \sum^m_{i=0} \epsilon_{ts, i}^2 \to \sigma_{ts}$. Thus, $\| x_{ts} \|^2 = \|\mu_{ts}\| + m \sigma_{ts}^2 $. 

In addition, although $ E[\|x_{ts}\|] = E[\sqrt{\|x_{ts}\|^2}] \leq \sqrt{E[\| x_{ts} \|^2]} $ by Jessen inequality, we show that $E[\|x_{ts}\|]$ can still be approximated as $\sqrt{E[\| x_{ts} \|^2]} = \sqrt{\|\mu_{ts}\| + m \sigma_{ts}^2}$ in high dimensional space. Since we know $ \|x_{ts}\|^2 = \|\mu_{ts}\|^2 + 2\mu_{ts}\epsilon_{ts} + \|\epsilon_{ts}\|^2 $, thus the variance $Var(\|x\|^2) = Var(2\mu_{ts}\epsilon_{ts})  + Var(\|\epsilon_{ts}\|^2) = 4\sigma^2_{ts}\|\mu_{ts}\|^2 + 2m\sigma_{ts}^4$. Therefore, the relative fluctuation of variable $ \| x_{ts} \|^2$ is :

\begin{align}
    \frac{std(\| x_{ts} \|^2)}{mean(\|x_{ts}\|^2)} &= \frac{\sqrt{4\sigma^2_{ts}\|\mu_{ts}\|^2 + 2m\sigma_{ts}^4}}{ \|\mu_{ts}\| + m \sigma_{ts}^2} \\
    &= \frac{\sqrt{4\sigma^2_{ts}\|\mu_{ts}\|^2} + \sqrt{2m}\sigma_{ts}^2}{ \|\mu_{ts}\| + m \sigma_{ts}^2 } \label{eq:relative_fluctuation}
\end{align}

In Eq. \ref{eq:relative_fluctuation}, the relative fluctuation of $ \| x_{ts} \|^2 \sim \frac{\sqrt{m}\sigma_{ts}^2}{m\sigma_{ts}} = \frac{1}{\sqrt{m}} $ and $ \frac{1}{\sqrt{m}} \in 0$ when $m$ is large (in high dimensional manifold / space), therefore, $\| x_{ts} \|^2$ will concentrate around its mean $ \|\mu_{ts}\| + m \sigma_{ts}^2 $. This makes $\|x_{ts}\| = \sqrt{\|x_{ts}\|^2} \approx \sqrt{E(\|x_{ts}\|^2)} = \sqrt{ \|\mu_{ts}\| + m \sigma_{ts}^2}$. Similarly, we apply the same process to language modality $x_l$ and obtain $\| x_l \| = \sqrt{\|x_{l}\|^2}\approx \sqrt{E(\|x_{l}\|^2)} = \sqrt{ \|\mu_{l}\| + n \sigma_{l}^2}$.

Theorem \ref{theorem_1:low_manifold_casues_pseudo_alignment} clearly indicates that low dimensionality of time series can cause the pseudo-alignment under the cone effect condition. To address this, one approach is to explicitly constrain the development of the cone effect within each LLM layer using a dedicated loss function \cite{gao2019representation}. However, this may inadvertently suppress the implicit multimodal alignment between time and language, potentially deactivating the knowledge transfer capabilities of the LLM. An alternative solution is to enhance the manifold dimensionality of time series data prior to LLM. This dimension enhancement should ideally raise the manifold dimension to similar level of language distribution, thereby enabling the model to better capture the internal structure of time series. This complex structure obstacle \say{easy route} but implicitly force LLM to distinguish the time series tokens, leading to mitigation of Pseudo-Alignment. Moreover, because the cone effect arises organically from the LLM’s normalization and attention mechanisms, the enhanced time‐series still benefit from this implicit alignment. In practice, the cone effect will continue to promote large angular separations and maintain cross‐modal coherence, thereby preserving both the model’s ability to align modalities and its capacity to capture fine‐grained unique feature of both modality.

\begin{figure*}[ht]
    \centering
    \includegraphics[width=\linewidth, trim=0 2em 0 0]{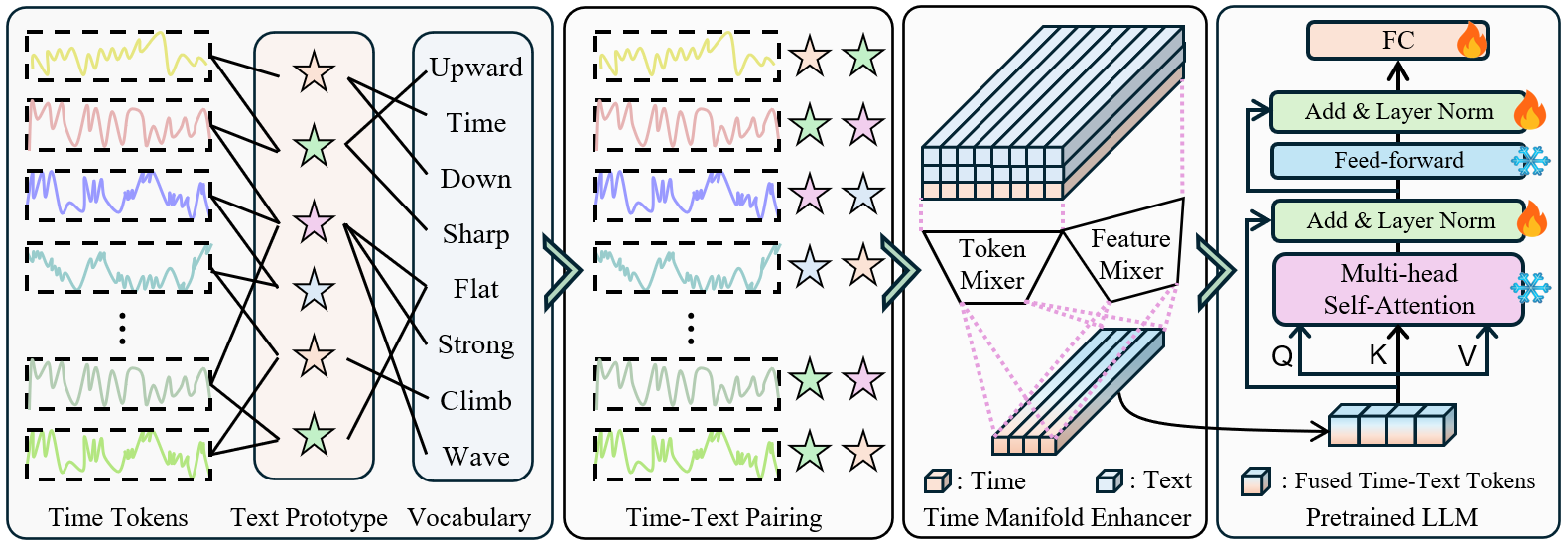}
    \caption{Overall Framework of Time Manifold Enhancer to address Pseudo-Alignment: (1) Text prototype is generated by a linear combination on the vocabulary and a one layer multihead attention is used to find text prototype best describing time series patches. (2) Each time patch will pick a top-k text prototype as the material to enhance dimension of time manifold. (3) Two MLP mixer are applied over the token dimension and feature dimension to compress and fuse the time-text tokens and generate the enhanced version of time tokens. (4) All fused time-text tokens will be sent to LLM for final prediction.}
    \label{fig:framework_overview}
\end{figure*}

\begin{table*}[ht]
    \centering
    \resizebox{\textwidth}{!}{%
    \begin{tabular}{cc|cc|cc|cc|cc|cc|cc|cc|cc|cc|cc}
         \toprule
         \multicolumn{2}{c|}{Dataset} & \multicolumn{2}{c|}{TimeSUP} & \multicolumn{2}{c|}{FSCA} & \multicolumn{2}{c|}{CALF} & \multicolumn{2}{c|}{S2IP} & \multicolumn{2}{c|}{TimeLLM} & \multicolumn{2}{c|}{UniTime} & \multicolumn{2}{c|}{OFA} & \multicolumn{2}{c|}{TimeMixer} & \multicolumn{2}{c|}{TimesNet} & \multicolumn{2}{c}{iTransformer} \\
         \multicolumn{2}{c|}{Pred Length} & MSE & MAE & MSE & MAE & MSE & MAE & MSE & MAE & MSE & MAE & MSE & MAE & MSE & MAE & MSE & MAE & MSE & MAE & MSE & MAE  \\
         \midrule
        
        \multirow{5}{*}{\begin{turn}{90} 
         ETTh1 
         \end{turn}} 
            & 96  & \textbf{0.376} & \textbf{0.400} & \underline{0.381} & \underline{0.404} & 0.386 & 0.410 & 0.394 & 0.415 & 0.383 & 0.412 & 0.397 & 0.418 & 0.392 & 0.407 & 0.388 & 0.402 & 0.384 & 0.402 & 0.386 & 0.405\\
            & 192 & \textbf{0.408} & \textbf{0.427} & \underline{0.416} & \underline{0.419} & 0.431 & 0.430 & 0.443 & 0.445 & 0.424 & 0.434 & 0.434 & 0.439 & 0.432 & 0.430 & 0.441 & 0.429 & 0.436 & 0.429 & 0.441 & 0.436  \\
            & 336 & \textbf{0.429} & \textbf{0.436} & 0.455 & 0.447 & 0.477 & 0.450 & 0.456 & 0.456 & \underline{0.437} & 0.453 & 0.468 & 0.457 & 0.441 & \underline{0.438} & 0.484 & 0.458 & 0.638 & 0.469 & 0.487 & 0.458 \\
            & 720 & \textbf{0.433} & \textbf{0.456} & \underline{0.455} & 0.471 & 0.473 & \underline{0.466} & 0.471 & 0.478 & 0.458 & 0.484 & 0.469 & 0.477 & 0.468 & 0.471 & 0.498 & 0.482 & 0.521 & 0.500 & 0.503 & 0.491 \\
            & Avg & \textbf{0.412} & \textbf{0.430} & \underline{0.426} & \underline{0.435} & 0.441 & 0.439 & 0.441 & 0.448 & \underline{0.426} & 0.445 & 0.442 & 0.448 & 0.433 & 0.436 & 0.452 & 0.442 & 0.495 & 0.450 & 0.454 & 0.447  \\
        \midrule

        \multirow{5}{*}{\begin{turn}{90} 
         ETTh2 
         \end{turn}} 
            & 96  & \textbf{0.293} & \textbf{0.346} & 0.300 & 0.351 & 0.297 & 0.351 & 0.300 & 0.354 & 0.295 & 0.353 & 0.295 & 0.345 & 0.311 & 0.366 & \underline{0.294} & 0.350 & 0.340 & 0.374 & 0.297 & \underline{0.349}  \\
            & 192 & \textbf{0.342} & \textbf{0.382} & \underline{0.348} & \underline{0.387} & 0.369 & 0.388 & 0.370 & 0.397 & 0.410 & 0.421 & 0.374 & 0.394 & 0.391 & 0.412 & 0.372 & 0.392 & 0.402 & 0.414 & 0.380 & 0.400  \\
            & 336 & \underline{0.369} & \underline{0.411} & \textbf{0.365} & \textbf{0.405} & 0.422 & 0.426 & 0.401 & 0.426 & 0.423 & 0.441 & 0.415 & 0.427 & 0.386 & 0.417 & 0.386 & 0.414 & 0.452 & 0.452 & 0.428 & 0.437  \\
            & 720 & \textbf{0.407} & \textbf{0.438} & \textbf{0.407} & \underline{0.444} & 0.421 & 0.450 & 0.428 & 0.455 & 0.455 & 0.467 & 0.425 & 0.444 & 0.433 & 0.456 & \underline{0.419} & 0.444 & 0.462 & 0.468 & 0.427 & 0.445  \\
            & Avg & \textbf{0.353} & \textbf{0.394} & \underline{0.355} & \underline{0.397} & 0.376 & 0.404 & 0.374 & 0.408 & 0.396 & 0.420 & 0.378 & 0.403 & 0.380 & 0.412 & 0.368 & 0.400 & 0.414 & 0.427 & 0.383 & 0.407  \\
        \midrule

        \multirow{5}{*}{\begin{turn}{90} 
         ETTm1 
         \end{turn}} 
            & 96  & \textbf{0.296} & \textbf{0.350} & \underline{0.301} & \underline{0.351} & 0.324 & 0.353 & 0.300 & 0.353 & 0.304 & 0.361 & 0.322 & 0.363 & 0.310 & 0.355 & 0.320 & 0.357 & 0.388 & 0.375 & 0.334 & 0.368  \\
            
            & 192 & \textbf{0.332} & \textbf{0.369} & \textbf{0.332} & \underline{0.371} & 0.369 & 0.376 & \underline{0.333} & 0.377 & 0.358 & 0.384 & 0.366 & 0.387 & 0.351 & 0.390 & 0.361 & 0.381 & 0.374 & 0.387 & 0.377 & 0.391 \\
            
            & 336 & \textbf{0.360} & \textbf{0.385} & 0.365 & \underline{0.391} & 0.407 & 0.399 & \underline{0.362} & 0.396 & 0.364 & 0.397 & 0.398 & 0.407 & 0.373 & 0.403 & 0.390 & 0.404 & 0.410 & 0.411 & 0.426 & 0.420 \\
            
            & 720 & \underline{0.411} & \textbf{0.418} & 0.421 & \underline{0.425} & 0.472 & 0.436 & \textbf{0.407} & 0.429 & 0.424 & 0.431 & 0.454 & 0.440 & 0.423 & 0.432 & 0.454 & 0.441 & 0.478 & 0.450 & 0.491 & 0.459 \\
            
            & Avg & \underline{0.350} & \textbf{0.380} & 0.354 & \underline{0.385} & 0.393 & 0.391 & \textbf{0.349} & 0.389 & 0.363 & 0.389 & 0.385 & 0.399 & 0.364 & 0.395 & 0.381 & 0.395 & 0.400 & 0.406 & 0.407 & 0.410 \\
        \midrule
            
        \multirow{5}{*}{\begin{turn}{90} 
         ETTm2 
         \end{turn}}
            & 96  & \underline{0.174} & 0.264 & \textbf{0.169} & 0.260 & 0.178 & \textbf{0.255} & 0.181 & 0.269 & 0.178 & 0.269 & 0.183 & 0.266 & 0.188 & 0.269 & 0.175 & \underline{0.258} & 0.187 & 0.267 & 0.180 & 0.264 \\
            
            & 192 & \textbf{0.233} & \underline{0.300} & \underline{0.237} & 0.306 & 0.243 & \textbf{0.298} & 0.257 & 0.323 & 0.241 & 0.313 & 0.251 & 0.310 & 0.251 & 0.309 & 0.251 & 0.309 & 0.251 & 0.312 & 0.249 & 0.309  \\
            
            & 336 & \underline{0.280} & 0.338 & \textbf{0.276} & \textbf{0.328} & 0.303 & 0.338 & 0.293 & 0.346 & 0.287 & \underline{0.337} & 0.319 & 0.351 & 0.307 & 0.346 & 0.298 & 0.340 & 0.321 & 0.351 & 0.311 & 0.348  \\
            
            & 720 & \textbf{0.358} & \textbf{0.381} & 0.366 & \underline{0.390} & 0.403 & 0.396 & 0.367 & 0.396 & \underline{0.364} & 0.401 & 0.420 & 0.410 & 0.426 & 0.417 & 0.391 & 0.396 & 0.408 & 0.403 & 0.412 & 0.407 \\
            
            & Avg & \textbf{0.261} & 0.323 & \underline{0.262} & \textbf{0.321} & 0.282 & \underline{0.322} & 0.279 & 0.331 & 0.267 & 0.330 & 0.293 & 0.334 & 0.293 & 0.335 & 0.275 & 0.325 & 0.291 & 0.333 & 0.288 & 0.332 \\
        \midrule
            
        \multirow{5}{*}{\begin{turn}{90} 
         Illness 
         \end{turn}} 
            & 24   & \underline{1.942} & \textbf{0.835} & 2.105 & 0.908 & 2.271 & 1.008 & 2.622 & 1.113 & \textbf{1.921} & \underline{0.902} & 2.346 & 0.954 & 2.205 & 0.982 & 2.241 & 0.986 & 2.317 & 0.934 & 2.347 & 1.731 \\
            & 36   & \underline{1.952} & 0.948 & 2.066 & 0.962 & 2.365 & 1.029 & 2.515 & 1.098 & 1.992 & 0.918 & 1.998 & \underline{0.912} & 2.166 & 0.969 & \textbf{1.863} & \textbf{0.903} & 1.972 & 0.920 & 2.468 & 0.998 \\
            & 48   & \textbf{1.738} & \textbf{0.886} & 1.902 & 0.906 & 2.395 & 1.046 & 2.530 & 1.066 & 2.187 & 1.031 & \underline{1.979} & \underline{0.912} & 2.220 & 0.963 & 2.402 & 1.088 & 2.238 & 0.940 & 2.489 & 1.016 \\
            & 60   & \textbf{1.907} & \underline{0.942} & 2.341 & 1.016 & 2.415 & 1.056 & 2.329 & 1.065 & 2.125 & 0.992 & 2.109 & 0.938 & 2.246 & 1.010 & 2.156 & 0.983 & \underline{2.027} & \textbf{0.928} & 2.471 & 1.065 \\
            & Avg  & \textbf{1.885} & \textbf{0.902} & 2.103 & 0.948 & 2.361 & 1.035 & 2.387 & 1.052 & \underline{2.056} & 0.961 & 2.108 & \underline{0.929} & 2.209 & 0.981 & 2.166 & 0.990 & 2.139 & 0.931 & 2.444 & 1.203 \\
        \midrule

        \multirow{5}{*}{\begin{turn}{90} 
        Weather 
         \end{turn}}
            & 96  & \textbf{0.151} & \textbf{0.200} & \underline{0.152} & 0.201 & 0.163 & 0.203 & 0.157 & 0.209 & 0.154 & \underline{0.199} & 0.171 & 0.214 & 0.157 & 0.208 & 0.163 & 0.209 & 0.172 & 0.220 & 0.174 & 0.214 \\
            & 192 & \textbf{0.197} & \textbf{0.243} & \textbf{0.197} & \underline{0.244} & 0.210 & 0.248 & 0.200 & 0.251 & 0.214 & 0.254 & 0.217 & 0.254 & \underline{0.200} & 0.246 & 0.208 & 0.250 & 0.219 & 0.261 & 0.221 & 0.254 \\
            & 336 & \textbf{0.248} & \textbf{0.284} & 0.253 & 0.289 & 0.266 & 0.290 & 0.253 & 0.288 & 0.273 & 0.291 & 0.274 & 0.293 & \underline{0.250} & 0.290 & 0.251 & \underline{0.287} & 0.280 & 0.306 & 0.278 & 0.296 \\
            & 720 & \textbf{0.330} & \underline{0.340} & \underline{0.332} & \underline{0.341} & 0.347 & 0.344 & 0.338 & \textbf{0.333} & 0.364 & 0.350 & 0.351 & 0.343 & 0.330 & 0.342 & 0.339 & 0.341 & 0.365 & 0.359 & 0.358 & 0.349 \\
            & Avg & \textbf{0.231} & \textbf{0.266} & \underline{0.233} & \underline{0.269} & 0.246 & 0.271 & 0.237 & 0.270 & 0.251 & 0.274 & 0.253 & 0.276 & 0.234 & 0.271 & 0.240 & 0.271 & 0.259 & 0.287 & 0.258 & 0.278 \\
        \midrule

        \multirow{5}{*}{\begin{turn}{90} 
         Traffic 
         \end{turn}}
            & 96  & \textbf{0.369} & 0.266 & \underline{0.372} & \textbf{0.255} & 0.433 & 0.292 & 0.379 & 0.274 & 0.388 & \underline{0.261} & 0.491 & 0.313 & 0.393 & 0.288 & 0.462 & 0.285 & 0.593 & 0.321 & 0.395 & 0.268 \\
            & 192 & \textbf{0.391} & \underline{0.277} & 0.399 & 0.284 & 0.451 & 0.297 & \underline{0.397} & 0.282 & 0.409 & \textbf{0.272} & 0.503 & 0.328 & 0.407 & 0.289 & 0.473 & 0.296 & 0.617 & 0.336 & 0.417 & 0.276 \\
            & 336 & \textbf{0.408} & \underline{0.287} & 0.413 & \textbf{0.273} & 0.467 & 0.302 & \underline{0.410} & 0.289 & 0.438 & 0.299 & 0.527 & 0.347 & 0.446 & 0.296 & 0.498 & 0.296 & 0.629 & 0.336 & 0.433 & 0.293 \\
            & 720 & \textbf{0.435} & \underline{0.298} & \underline{0.440} & \textbf{0.291} & 0.485 & 0.323 & 0.445 & 0.301 & 0.462 & 0.301 & 0.550 & 0.371 & 0.469 & 0.312 & 0.506 & 0.313 & 0.640 & 0.350 & 0.467 & 0.302 \\
            & Avg & \textbf{0.400} & \underline{0.282} & \underline{0.406} & \textbf{0.276} & 0.459 & 0.303 & 0.407 & 0.287 & 0.424 & 0.283 & 0.518 & 0.340 & 0.428 & 0.296 & 0.484 & 0.297 & 0.620 & 0.336 & 0.428 & 0.284 \\
        \midrule

        \multirow{5}{*}{\begin{turn}{90}
         ECL 
         \end{turn}}
            & 96  & \textbf{0.128} & \textbf{0.221} & 0.133 & 0.228 & 0.167 & 0.254 & \underline{0.131} & \underline{0.226} & 0.172 & 0.265 & 0.196 & 0.287 & 0.137 & 0.236 & 0.153 & 0.247 & 0.168 & 0.272 & 0.148 & 0.240 \\
            & 192 & \textbf{0.151} & \textbf{0.247} & 0.157 & \underline{0.250} & 0.178 & 0.264 & 0.158 & 0.252 & 0.182 & 0.279 & 0.199 & 0.291 & \underline{0.153} & 0.253 & 0.166 & 0.256 & 0.184 & 0.289 & 0.162 & 0.253 \\
            & 336 & \textbf{0.170} & \textbf{0.269} & \underline{0.176} & 0.281 & 0.196 & 0.282 & 0.183 & 0.280 & 0.195 & 0.288 & 0.214 & 0.305 & 0.183 & 0.280 & 0.185 & \underline{0.277} & 0.198 & 0.330 & 0.178 & \textbf{0.269} \\
            & 720 & \textbf{0.207} & \textbf{0.299} & 0.215 & \underline{0.302} & 0.232 & 0.312 & 0.210 & 0.306 & 0.233 & 0.320 & 0.254 & 0.335 & \underline{0.209} & 0.308 & 0.225 & 0.310 & 0.220 & 0.320 & 0.225 & 0.317 \\
            & Avg & \textbf{0.164} & \textbf{0.256} & \underline{0.170} & \underline{0.265} & 0.193 & 0.278 & \underline{0.170} & 0.266 & 0.195 & 0.288 & 0.216 & 0.304 & \underline{0.170} & 0.269 & 0.182 & 0.272 & 0.192 & 0.295 & 0.178 & 0.270 \\
            \midrule
         \multicolumn{2}{c|}{Wins} & \multicolumn{2}{c|}{ 60 } & \multicolumn{2}{c|}{ 13 } & \multicolumn{2}{c|}{ 2 } & \multicolumn{2}{c|}{ 2 } & \multicolumn{2}{c|}{ 2 } & \multicolumn{2}{c|}{ 0 } & \multicolumn{2}{c|}{ 0 } & \multicolumn{2}{c|}{ 2 } & \multicolumn{2}{c|}{ 1 } & \multicolumn{2}{c}{ 0 }\\
         \bottomrule
    \end{tabular}
    }
    \caption{Full Results Long-term Forecasting with LLM4TS: \textbf{Bold Text} indicate the best performer and \underline{Underline Text} indicate the second best performer. We count the number of wins by the number of best performer.}
    \label{tab:LLM4TS_results_AP}
\end{table*}

\section{Address Pseudo-Alignment}
To this end, we proposed TimeSUP to address Pseudo-Alignment problem by increasing data manifold dimension of time series data to help language model to understand time series tokens. As shown in Figure \ref{fig:framework_overview}, our proposed method follows the analysis from the preliminary and aims to achieve: (1) Find most relevant material as enhanced fusion material and (2) manifold dimension enhancement to mitigate Pseudo-Alignment. We expect the well-alignment should preserve distinct feature and remain high similarity.

\noindent \textbf{Patch Time Series Embedding.} Each input series $x_{ts}$ with sequence length $L$ will be first normalized to zero mean and unit standard deviation by reversible normalization layer. Then, it will be patched into several patches with patch size $P$ and stride step $S$ \cite{nie2022time}, producing number of patch $N = \lceil \frac{P - L}{S} \rceil + 1 $. The patched time series patches will be linearly spanned to the shared language embedding space $\mathbb{R}^d$ to produce time series token $T_i \in \mathbb{R}^d$, where $d$ is the hidden size of LLM backbone. 

\noindent \textbf{Top-K Text Prototype.} Each time series patch contains strong temporal and localized pattern and we hypothesis that proper selection of language token can describe each time series patch (e.g. time is climbing). To help LLM understand the time series tokens from semantic level, we adapt text prototype \cite{jin2023time} and find the most relevant language prototype to time series patches, where each time series token will be paired with Top-K language prototype to best describe the semantic feature about the time patch. Specifically, the text prototype $L^*_i \in \mathbb{R}^d$ is generated by linear combination of the entire vocabulary as illustrate in Figure \ref{fig:framework_overview}. The number of prototype is set to 1000 by default. 

Then, an asymmetric cross-attention is employed to find Top-K prototype that best describe for each time patch. We defined the query matrix $\mathbf{Q} = \mathbf{T} \mathbf{W}_q $ as the time series token $\mathbf{T} \in \mathbb{R}^{N \times d}$ and key matrix $\mathbf{K} = \mathbf{L}^* \mathbf{W}_k $ as prototype, where $\mathbf{L}^* \in \mathbb{R}^{1000 \times d}$,  $\mathbf{W}_k, \mathbf{W}_q \in \mathbb{R}^{d}$ and $\mathbf{Q}, \mathbf{K} \in \mathbb{R}^{d}$. We calculate the attention weight by the dot-product of $\mathbf{Q}$ and $\mathbf{K}$ to obtain Top-K attention weight $\mathbf{A}_k \in \mathbb{R}^{k \times N}$ as shown in Eq. \ref{eq:attention}. Note that we consider $\mathbf{A}_k$ to retrieve the TopK index to select the most relevant text prototype $\mathbf{L}^*_K \in \mathbb{R}^{K \times N \times d}$ from $\mathbf{L}^*$ to each corresponding time token. 

\begin{equation}
    \mathbf{A}_k = TopK\left(Softmax(\frac{\mathbf{QK^{\top}}}{\sqrt{d}})\right) \label{eq:attention}
\end{equation}

\begin{figure*}[ht]
    \centering
    \includegraphics[width=\linewidth,trim=0 3em 0 0]{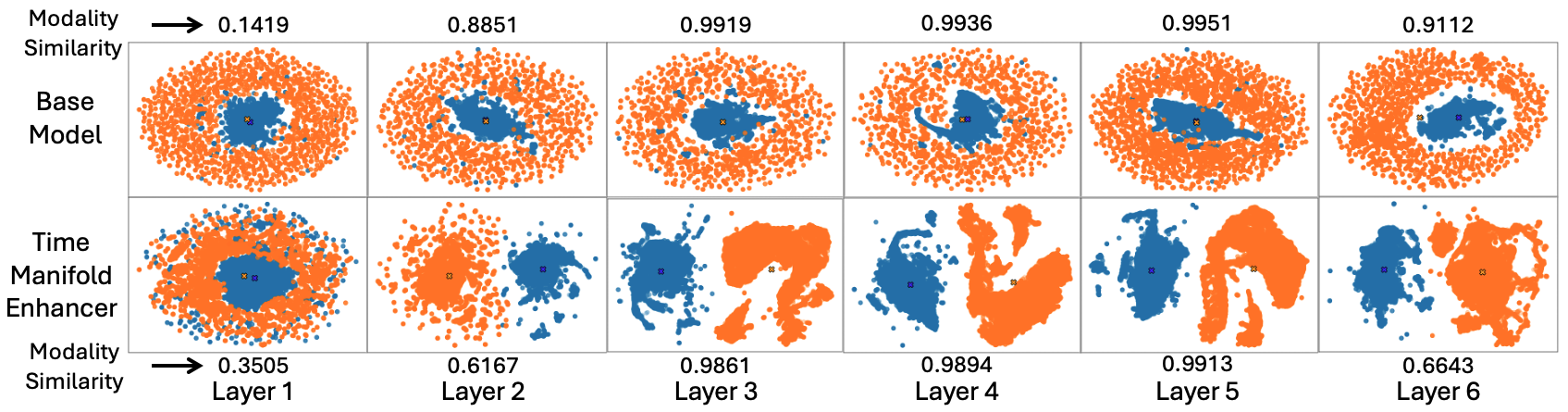}
    \caption{Representation in Layers: the cosine similarity is calculated between time and language tokens after each layer, \textcolor{blue}{Blue} is language and \textcolor{orange}{Orange} is time series. LLM starts to recognize time series token after layer 2 as two modalities are separable while maintain high cosine similarity.}
    \label{fig:alignment_in_LLLM_Layers}
\end{figure*}

\noindent \textbf{Time Manifold Enhancer.} Motivated by the previous analysis in which the time series lies in lower data manifold than language, The TopK prototypes are considered as a meaningful manifold enhancement material and a multimodal fusion with complex data and simple data will eventually produce a higher manifold dimension. Then,we leverage the Top‑K language prototypes to enrich each time series token’s manifold dimension, bringing its manifold dimensionality closer to that of language tokens. We designed two lightweight MLPs, $\mathbf{M}_{c} \in \mathbb{R}^{((K+1)\times N) \times n}$ operating across token dimension and $\mathbf{M}_{f} \in \mathbb{R}^{d \times d}$ across feature channels to fuse each time token with its selected prototype. Specifically, $\mathbf{L}^*_K$ will be concatenated with $\mathbf{T}$ to form a time-text pairs $\mathbf{T}_t \in \mathbb{R}^{((K + 1) \times N) \times d}$. We then compress $\mathbf{T}_t$ into a fixed sequence length $n$ (e.g. 8 or 16) fused representation $\mathbf{T}^* \in \mathbb{R}^{n \times d}$ by $\mathbf{M}_c$. Next, the fused representation is then and fed into $\mathbf{M}_f$ for further feature fusion as shown in Eq. \ref{eq:enhancer}.

\begin{equation}
    \mathbf{T}^* = \mathbf{M}_f (\mathbf{M}_c^{\top} \mathbf{T}_t)^{\top} \label{eq:enhancer}
\end{equation}

This multimodal fusion substantially elevates the intrinsic dimensionality of the time series manifold: under a PCA probe, the number of time series principle component increases from 21 to 224 (compared to 712 for language tokens in GPT‑2) as shown in Figure \ref{fig:PCA_EVR}.  We find this is particularly useful for not only manifold enhancement, but also improving the speed for LLM forward as the number of tokens are constrained to a certain number. By grafting the high‑dimensional structure of language onto time‑series data, our time enhancer produces richer token embeddings that preserve modality‑specific distinctions while maintaining strong alignment on shared semantic features, ultimately boosting downstream performance and activating pretrained language knowledge and avoid pseudo-alignment problem.

\section{Related Work}
\subsection{Lightweight Time Series Models}
Most of time series forecasting approaches are designed specific tasks and trained from scratch. Those time series models focus on modeling of temporal relationship of time series and has been widely explored. Typically, we can categorized the time series model into the following architecture: RNN-, CNN-, Transformer- and MLP-based methods. RNN-based models learned temporal feature by recursively temporal hidden states \cite{chen2020amrnn,chen2019damtrnn,wen2017multi}. CNN-based model, on the other hand, learn a series of kernels to slide over the time series and extract temporal and localized feature \cite{ismail2020inceptiontime}. However, RNN- and CNN-based models usually suffer from the long-term dependency as the architecture focus more on local pattern. Therefore, recent works take advantages of global receptive field of Transformer for long-term feature extraction such as PatchTST \cite{nie2022time}, \textbf{iTransformer}\cite{liu2023itransformer}, FedFormer\cite{zhou2022fedformer}, AutoFormer\cite{wu2021autoformer}. Furthermore, some researches argue that MLP-based method (e.g. DLinear\cite{zeng2023transformers} \textbf{TimesNet} \cite{wu2022timesnet}, \textbf{TimeMixer} \cite{wang2024timemixer}, FITS \cite{xu2023fits}) can even outperform transformer-based models and significantly simplified the parameter size. (\textbf{Bold Text} indicate the selected baselines in experiments)

\subsection{Large Language Models for Time Series}
Recently, large language models have been proved to have capability on strong generalization to time series analysis. \citet{gruver2023large} convert time series as a series of digit token for zero-shot forecasting. \textbf{OFA} \citet{zhou2023one} reprogrammed GPT-2 to adapt time series multi-tasks by finetuning the LayerNorm layer. \textbf{TimeLLM} \cite{jin2023time} employed prompts and cross attention to find the text tokens to best describe temporal feature from vocabulary. \textbf{CAFL}\cite{liu2025calf} leverages LoRa finetuning and textual-temporal consistent loss to achieve model reprogramming. \textbf{S2IP} \cite{pan2024textbf} decompose time series and align the language tokens to STL components. \textbf{UniTime} \cite{liu2024unitime} attempt to train a unifying forecaster for time series by dataset description prompts and temporal feature of time. TimeCMA \cite{liu2025timecma} trained an individual time series encoder alongside a pretrained LLM for language-time similarity retrieval. \textbf{FSCA} \cite{hu2025context} introduce GNN to generate time-text interaction between time and text modality. (\textbf{Bold Text} indicate the selected baselines in experiments)

\section{Experiment}

\noindent\textbf{Set-up} \ We evaluated TimeSUP on widely used benchmarking dataset for long-term forecasting, ETT(4 subsets), Illness, Weather, Traffic and Electricity \cite{zheng2024revisited, zhou2023one} for comprehensive comparison. All baseline are reproduced based on the best parameters from the original paper or github repository. The evaluation metrics include mean square error (MSE) and mean absolute error(MAE). Our ablation study is implemented on ETTh1 with 720 prediction length. Each Experiment were evaluate on 4 $\times$ RTX 4090 24GB and 4 $\times$ A100 40GB and default random seed from OFA \cite{zhou2023one}. All details parameter set-up for our proposed method is shown in Appendix. We will release our github repository upon the acceptance of paper.

In addition, we also visualize time and language tokens in layers of base model and base model with TimeSUP to show the impact of TimeSUP shown in Figure \ref{fig:alignment_in_LLLM_Layers}. Moreover, we implement two additional ablation experiment to investigate the adaptability of Time Enhancer module in TimeSUP and what component in LLM causes Pseudo-Alignment as shown in Table \ref{tab:adapt_enhancer}.

\begin{table*}[ht]
    \centering
    \resizebox{\textwidth}{!}{%
    \setlength{\tabcolsep}{0.85mm}
    \begin{tabular}{cc|cc|cc||cc|cc||cc|cc||cc|cc||cc|cc}
         \toprule

         \multicolumn{2}{c|}{Model}     & \multicolumn{4}{c||}{CALF} & \multicolumn{4}{c||}{S2IP} & \multicolumn{4}{c||}{TimeLLM} & \multicolumn{4}{c||}{UniTime} & \multicolumn{4}{c}{OFA} \\
         \midrule 
         \multicolumn{2}{c|}{Enhancer?}  & \multicolumn{2}{c}{\ding{51}} & \multicolumn{2}{c||}{\ding{55}}  & \multicolumn{2}{c}{\ding{51}} & \multicolumn{2}{c||}{\ding{55}} & \multicolumn{2}{c}{\ding{51}} & \multicolumn{2}{c||}{\ding{55}} & \multicolumn{2}{c}{\ding{51}} & \multicolumn{2}{c||}{\ding{55}} & \multicolumn{2}{c}{\ding{51}} & \multicolumn{2}{c}{\ding{55}} \\
         \midrule 
         \multicolumn{2}{c|}{Pred Length}  & MSE & MAE & MSE & MAE & MSE & MAE & MSE & MAE & MSE & MAE & MSE & MAE & MSE & MAE & MSE & MAE & MSE & MAE & MSE & MAE \\
         \midrule
        
        \multirow{5}{*}{\begin{turn}{90} 
         ETTh1 
         \end{turn}} 
            & 96  & \textbf{0.386} & \textbf{0.397} & 0.386 & 0.410 & \textbf{0.383} & \textbf{0.413} & 0.394 & 0.415 & 0.388 &  0.414 & \textbf{0.383} & \textbf{0.412} & \textbf{0.384} & \textbf{0.410} & 0.397 & 0.418 & \textbf{0.367} & \textbf{0.400} & 0.392 & 0.407 \\
            
            & 192 & \textbf{0.427} & \textbf{0.422} & 0.431 & 0.430 & \textbf{0.408} & \textbf{0.428} & 0.433 & 0.445 & \textbf{0.412} & \textbf{0.433} & 0.424 & 0.434 & \textbf{0.430} & \textbf{0.435} & 0.434 & 0.439 & \textbf{0.408} & \textbf{0.427} & 0.432 & 0.430  \\
            
            & 336 & \textbf{0.473} & \textbf{0.449} & 0.477 & 0.450 & \textbf{0.422} & \textbf{0.442} & 0.456 & 0.456 & \textbf{0.428} & \textbf{0.447} & 0.437 & 0.453 & 0.470 & 0.447 & \textbf{0.468} & \textbf{0.457} & \textbf{0.429} & \textbf{0.436} & 0.441 & 0.438  \\
            
            & 720 & \textbf{0.468} & \textbf{0.465} & 0.473 & 0.466 & \textbf{0.457} & \textbf{0.474} & 0.471 & 0.478 & \textbf{0.453} & \textbf{0.478} & 0.458 & 0.484 & \textbf{0.447} & \textbf{0.461} & 0.469 & 0.477 & \textbf{0.433} & \textbf{0.456} & 0.468 & 0.471  \\
            
            & Avg & \textbf{0.439} & \textbf{0.433} & 0.441 & 0.439 & \textbf{0.418} & \textbf{0.439} & 0.441 & 0.448 & \textbf{0.420} & \textbf{0.443} & 0.426 & 0.445 & \textbf{0.432} & \textbf{0.438} & 0.442 & 0.448 & \textbf{0.412} & \textbf{0.430} & 0.433 & 0.436  \\
            \midrule

        \multirow{5}{*}{\begin{turn}{90} 
        Illness 
         \end{turn}} 
            & 24  & \textbf{2.226} & \textbf{0.961} & 2.271 & 1.008 & \textbf{2.136} & \textbf{0.979} & 2.622 & 1.113 & \textbf{1.867} & \textbf{0.851} & 1.921 & 0.902 & \textbf{2.306} & \textbf{0.934} & 2.346 & 0.954 & \textbf{1.942} & \textbf{0.835} & 2.205 & 0.982 \\
            & 36  & \textbf{1.831} & \textbf{0.848} & 2.365 & 1.029 & \textbf{2.183} & \textbf{0.990} & 2.515 & 1.098 & \textbf{1.883} & \textbf{0.879} & 1.992 & 0.918 & 2.003 & 0.922 & \textbf{1.998} & \textbf{0.912} & \textbf{1.952} & \textbf{0.948} & 2.166 & 0.969 \\
            & 48  & \textbf{1.815} & \textbf{0.860} & 2.395 & 1.046 & \textbf{2.158} & \textbf{0.976} & 2.530 & 1.066 & \textbf{1.789} & \textbf{0.840} & 2.187 & 1.031 & 2.139 & \textbf{0.923} & \textbf{1.979} & 0.912 & \textbf{1.738} & \textbf{0.886} & 2.220 & 0.963 \\
            & 60  & \textbf{1.892} & \textbf{0.879} & 2.415 & 1.056 & \textbf{2.085} & \textbf{1.009} & 2.329 & 1.065 & \textbf{1.852} & \textbf{0.875} & 2.125 & 0.992 & \textbf{2.073} & \textbf{0.934} & 2.109 & 0.938 & \textbf{1.907} & \textbf{0.942} & 2.246 & 1.010 \\
            & Avg & \textbf{1.941} & \textbf{0.887} & 2.361 & 1.035 & \textbf{2.141} & \textbf{0.989} & 2.387 & 1.052 & \textbf{1.848} & \textbf{0.861} & 2.056 & 0.961 & 2.130 & \textbf{0.928} & \textbf{2.108} & 0.929 & \textbf{1.885} & \textbf{0.902} & 2.209 & 0.981 \\
         \bottomrule
    \end{tabular}
·    }
    \caption{Comparison Baselines with and without Manifold Enhancer: \textbf{Bold} indicate the improvement over original model}
    \label{tab:adapt_enhancer}
\end{table*}

\begin{figure*}[ht]
    \centering
    \includegraphics[width=\linewidth, trim=0 3em 0 0]{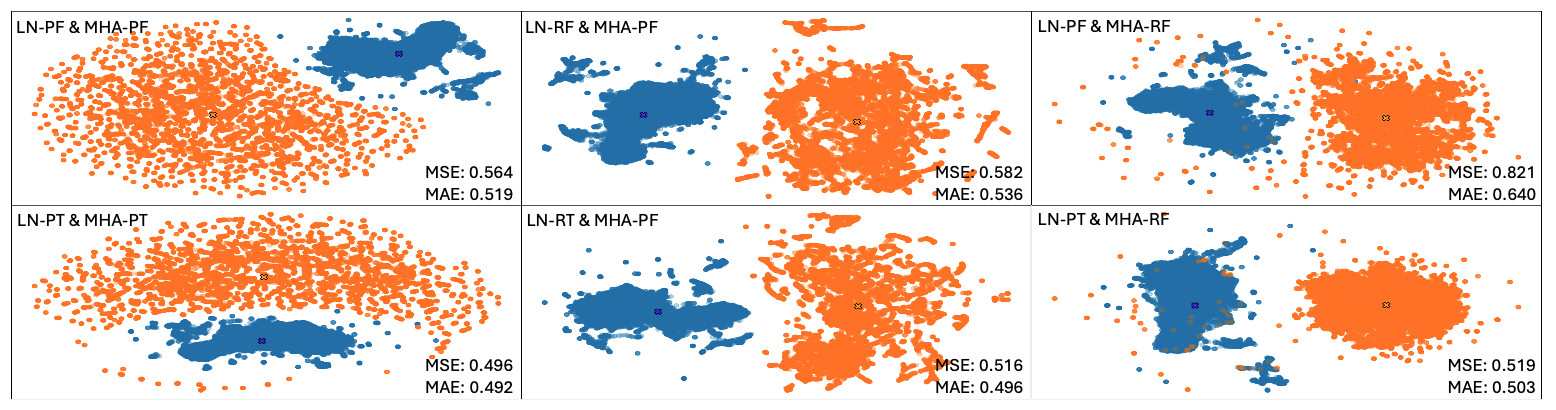}
    \caption{UMAP Representation for Different Setting on ETTh1-720 dataset. \textcolor{blue}{Blue} is language and \textcolor{orange}{Orange} is time series. Notation: LayerNorm(LN), MultiHead Attention(MHA), Pretrained–Frozen (PF), Pretrained–fineTuned (PT), Random–Frozen (RF), and Random–fineTuned (RT).}
    \label{fig:UMAP_Repre_for_alignment}
\end{figure*}

\subsection{Primary Results} 
As shown in Table \ref{tab:LLM4TS_results_AP}, our proposed method outperforms other LLM4TS and typical lightweight model in 8 long-term forecasting dataset. In particular, our proposed method wins 60/80 on acorss baseline models respectively. Moreover, those LLM4TS baseline can not consistently wins over lightweight model in Weather and ECL due to the harm of Pseudo-Alignment problem in those LLM4TS approach. TimeSUP, on the other hand, addresses Pseudo-Alignment effectively by increasing the dimension of time manifold, leading to superior improvement over LLM4TS and significant improvement over lightweight models. Although other LLM4TS approaches, such as S2IP and TimeLLM, are competitive to TimeSUP in ETTm1 and ETTm2, TimeSUP still win them consistently on ETTh1, ETTh2 and other large scale dataset. We show more detailed results with standard deviation in Appendix Table \ref{tab:LLM4TS_with_std} that shows TimeSUP is stabler than other LLM4TS method.

\subsection{Pseudo-Alignment in LLM Layers}
To investigate the progression of pseudo‐alignment across the LLM’s depth, we plot UMAP embeddings of language and time series tokens alongside their layer‐wise cosine similarity (Figure \ref{fig:alignment_in_LLLM_Layers}) by implementing visualization on ETTh1-720 and 6 layer GPT-2. The base model is GPT-2 fed only with time series tokens and finetuning layernorm layer. In the base model, cosine similarity surges to nearly 1 by the first layer and remains above 0.9 through all subsequent layers, indicating an almost complete collapse of the two modalities. Indeed, both the base and our proposed methods exhibit this centroid collapse at layer, where language and time series tokens become indistinguishable. However, from layer 2 onward, our approach diverges: the UMAP projections show that time series embeddings begin to fan out and map onto the language manifold without full collapse, and the cosine similarity, while still rising. This pattern demonstrates that our method successfully mitigates pseudo‐alignment, preserving time series internal structure, while retaining enough implicit cone effect alignment to support cross‑modal coherence. In addition, the increase in cosine similarity of TimeSUP proceeds more gradually than in the base model, and the two modalities remain clearly separable, indicating that the model actively distinguishes time‑series tokens rather than collapsing them into the language centroid for a \say{perfect} alignment. As the network deepens, TimeSUP language and time representations gradually settle into their own distinct cones. Although their cosine similarity continues to rise, it plateaus at approximately 0.6643 by the final layer. This behavior is exactly what one would expect from a well‑generalized multimodal model: it preserves each modality’s unique structure while maintaining substantial cross modal coherence, in line with the analysis in preliminary section. This explains that the high similarity in hidden layer from TimeSUP come from the actual alignment between tokens instead of the distribution centroid point.

\subsection{Is the Enhancer Useful for other LLM4TS Model?}
By integrating our TimeSUP enhancer into several leading LLM4TS methods—namely CALF, S2IP, UniTime and TimeLLM, we ensure that only minimal modifications are applied: the Time Manifold Enhancer is inserted immediately before the LLM backbone, and, where applicable, the existing text‐prototype selection mechanism remains unchanged. OFA serves as the underlying backbone for TimeSUP, so no additional alterations are needed. As reported in Table \ref{tab:adapt_enhancer}, this straightforward adaptation yields an average reduction in MSE of approximately 11\% on the Illness dataset and 2\% on ETTh1 across all tested configurations. The most pronounced gains appear in OFA and S2IP: on ETTh1, S2IP+TimeSUP achieves a 3\% lower MSE and a 2\% lower MAE compared to vanilla S2IP, while OFA+TimeSUP obtains a 4.8\% reduction in MSE and a 1.3\% reduction in MAE relative to OFA. This improvement is also consistent in Illness dataset. These improvements largely reflect the effectiveness of each model’s text‐prototype selection when coupled with our enhancer. Conversely, TimeLLM shows the smallest performance increase on both ETTh1 and Illness; we attribute this to the relatively large number of prompt tokens in TimeLLM, many of which may be redundant for representing local temporal structure despite their utility for reasoning. Notably, the most competitive approach, FSCA \cite{hu2025context}, implicitly follows a similar idea to TimeSUP by enhancing the dimensionality of time series tokens due to the GNN information passing, although this is not explicitly discussed in their work. However, adapting TimeSUP within FSCA’s framework required substantial modifications to their implementation; therefore, we exclude FSCA from this comparison.

\subsection{What Components in LLM Manipulate the Feature Map of Time Series?}
This section empirically investigates how pretrained LayerNorm and Multi-Head Attention (MHA) contribute to Pseudo-Alignment of time series tokens. Building on our hypothesis that these components \say{take easy route} manipulate time-series embeddings to induce Pseudo-Alignment, we conduct an ablation study on a 6-layer GPT-2, as illustrated in Figure \ref{fig:UMAP_Repre_for_alignment}. We examine representation of time and language tokens under four conditions for combination setting: Pretrained–Frozen (PF), Pretrained–fineTuned (PT), Random–Frozen (RF), and Random–fineTuned (RT). This experiment will reveal how the pretrained language knowledge in multihead attention and layernorm layer impact Pseudo-Alignment and forecasting performance. Across all models with randomly initialized LN or MHA, we observe a pronounced separation between time series and language embeddings. By contrast, any model incorporating pretrained LN or MHA exhibits a strong tendency toward pseudo‑alignment, replicating the behavior shown in Figure \ref{fig:alignment_in_LLLM_Layers}. In particular, the combination of LN‑PT and MHA‑PT produces the most severe pseudo‑alignment: these pretrained components reshape the time series feature manifold toward the language domain, while LN‑PT dynamically modulates this shift to facilitate manipulation by the MHA‑PF layers, as in OFA. Conversely, fine‑tuning the MHA layers (MHA‑FT) further perturbs the model’s language representations by entangling them with the time series signal, thereby reducing pseudo‑alignment. Moreover, any random initialization within these components significantly degrades forecasting performance: LN‑PF \& MHA‑RF yields the greatest performance drop, whereas LN‑RT \& MHA‑PF incurs the smallest, indicating that most language knowledge is preserved in the MHA layer. Together, these findings demonstrate that the interaction between pretrained LN and MHA is the principal driver of pseudo‑alignment, with LN‑PT \& MHA‑PF causing the worst pseudo‑alignment, as shown in Figure \ref{fig:no-mixer}.

\section{Conclusion}
In this work, we close the gap that limited insights and solution regarding pseudo-alignemnt problem by implementing comprehensive empirical analysis and proposing an effective \say{plug-and-play} time enhancer module. We showed that Pseudo-Alignment problem is a synthetic effect of cone effect and low data manifold of time series data, building up the bridge of cone effect in LLM to time series analysis. With inspiration from the analysis, we proposed TimeSUP module to effectively mitigate this issue and consistently outperforms other LLM- and lightweight-based time series forecaster.

\bibliographystyle{ACM-Reference-Format}
\bibliography{sample-base}

\appendix
\newpage
\onecolumn        
\section{Additional Experiment Details}  \label{ap:additional_experiments}

\begin{table*}[ht]
\centering
\setlength{\tabcolsep}{0.85mm}
\caption{Dataset statistics are from \cite{jin2023time}. The dimension indicates the number of time series channel and the size is organized as (training, validation, testing).} 
\label{tab:dataset_stats}
\begin{tabular}{c|ccccc}
    \toprule
    Dataset & Dim. & Prediction Length & Dataset Size & Frequency & Domain \\
    \midrule
    ETTh1      & 7 & {96, 192, 336, 720} & (8545, 2881, 2881) & 1 hour & Temperature \\
    ETTh2      & 7 & {96, 192, 336, 720} & (8545, 2881, 2881) & 1 hour & Temperature \\
    ETTm1      & 7 & {96, 192, 336, 720} & (34465, 11521, 11521) & 15 mins & Temperature \\
    ETTm2      & 7 & {96, 192, 336, 720} & (34465, 11521, 11521) & 15 mins & Temperature \\
    Illness    & 7 & {24, 36, 48, 60} & (617, 74, 170) & 1 week & Healthcare \\
    Weather    & 21 & {96, 192, 336, 720} & (36792, 5271, 10540) & 10 mins & Weather \\
    Traffic    & 862 & {96, 192, 336, 720} & (12185, 1757, 3509) & 1 hour & Transportation \\
    ECL        & 862 & {96, 192, 336, 720} & (18317, 2633, 5261) & 1 hour & Electricity \\
    \bottomrule
\end{tabular}
\end{table*}

The configurations of our model various to datasets while most configurations are consistent. By default, the Adam optimizer and Mean Square Error loss function are applied to trained the proposed method. The parameter described in Table \ref{tab:parameter} are the best parameter to reproduce the main results in Table \ref{tab:LLM4TS_results_AP}.

\begin{table*}[ht]
\centering
\caption{Hyperparameter Setting for results reproduction}
\label{tab:parameter}
\begin{tabular}{c|cccccccc}
    \toprule
    Dataset & Input Length & Patch Size & Stride & Dropout & Learn. Rate & Comp. Token & Top-k Prototype & \# LLM layer \\
    \midrule
    ETTh1      & 336 & 16 & 8 & 0.3 & 1e-4 & 10 & 4 & 6 \\
    ETTh2      & 512 & 16 & 8 & 0.2 & 5e-4 & 32 & 4 & 6 \\
    ETTm1      & 512 & 16 & 16 & 0.2 & 1e-4 & 32 & 4 & 6 \\
    ETTm2      & 512 & 16 & 16 & 0.3 & 1e-4 & 32 & 6 & 6 \\
    Illness    & 104 & 24 & 2 & 0 & 1e-4 & 8 & 4 & 6 \\
    Weather    & 336 & 16 & 8 & 0.1 & 1e-4 & 16 & 4 &  6\\
    Traffic    & 336 & 16 & 8 & 0 & 1e-3 & 10 & 4 & 6 \\
    ECL        & 336 & 16 & 8 & 0.1 & 1e-4 & 16 & 4 & 6 \\
    \bottomrule
\end{tabular}
\end{table*}

We found that FCSA, CALF S2IP and TimeLLM are the competitive model we have in the table. Therefore, to show the effectiveness of our method, we show resutls with variance in Table \ref{tab:LLM4TS_with_std} under 3 different random seeds, which is the same source of results as Table \ref{tab:LLM4TS_results_AP}, but with standard deviation to show that it is actually outperform others statistically.

\begin{table*}[ht]
    \centering
     \resizebox{\textwidth}{!}{%
    \begin{tabular}{cc|cc|cc|cc|cc|cc}
         \toprule
         \multicolumn{2}{c|}{Dataset} & \multicolumn{2}{c|}{TimeSUP} & \multicolumn{2}{c|}{FSCA} & \multicolumn{2}{c|}{CALF} & \multicolumn{2}{c|}{S2IP} & \multicolumn{2}{c}{TimeLLM} \\
         \multicolumn{2}{c|}{Pred Length} & MSE & MAE & MSE & MAE & MSE & MAE & MSE & MAE & MSE & MAE  \\
         \midrule
        
        \multirow{5}{*}{\begin{turn}{90} 
         ETTh1 
         \end{turn}} 
            & 96  & \textbf{0.376} {\scriptsize $\pm$ $6e^{-3}$} & \textbf{0.400} {\scriptsize $\pm$ $4e^{-3}$} & \underline{0.381} {\scriptsize $\pm$ $4e^{-3}$} & \underline{0.404} {\scriptsize $\pm$ $7e^{-3}$} & 0.386 {\scriptsize $\pm$ $4e^{-3}$} & 0.410 {\scriptsize $\pm$ $8e^{-3}$} & 0.394 {\scriptsize $\pm$ $11e^{-3}$} & 0.415 {\scriptsize $\pm$ $10e^{-3}$} & 0.383 {\scriptsize $\pm$ $10e^{-3}$} & 0.412 {\scriptsize $\pm$ $8e^{-3}$}  \\
            & 192 & \textbf{0.408} {\scriptsize $\pm$ $7e^{-3}$} & \textbf{0.427} {\scriptsize $\pm$ $10e^{-3}$} & \underline{0.416} {\scriptsize $\pm$ $12e^{-3}$} & \underline{0.419} {\scriptsize $\pm$ $11e^{-3}$} & 0.431 {\scriptsize $\pm$ $18e^{-3}$} & 0.430 {\scriptsize $\pm$ $14e^{-3}$} & 0.443 {\scriptsize $\pm$ $8e^{-3}$} & 0.445 {\scriptsize $\pm$ $8e^{-3}$} & 0.424 {\scriptsize $\pm$ $6e^{-3}$} & 0.434 {\scriptsize $\pm$ $15e^{-3}$} \\
            & 336 & \textbf{0.429} {\scriptsize $\pm$ $12e^{-3}$} & \textbf{0.436} {\scriptsize $\pm$ $10e^{-3}$} & 0.455 {\scriptsize $\pm$ $7e^{-3}$} & 0.447 {\scriptsize $\pm$ $11e^{-3}$} & 0.477 {\scriptsize $\pm$ $6e^{-3}$} & 0.450 {\scriptsize $\pm$ $6e^{-3}$} & 0.456 {\scriptsize $\pm$ $7e^{-3}$} & 0.456 {\scriptsize $\pm$ $13e^{-3}$} & \underline{0.437} {\scriptsize $\pm$ $11e^{-3}$} & 0.453 {\scriptsize $\pm$ $13e^{-3}$} \\
            & 720 & \textbf{0.433} {\scriptsize $\pm$ $2e^{-3}$} & \textbf{0.456} {\scriptsize $\pm$ $3e^{-3}$} & \underline{0.455} {\scriptsize $\pm$ $4e^{-3}$} & 0.471 {\scriptsize $\pm$ $4e^{-3}$} & 0.473 {\scriptsize $\pm$ $3e^{-3}$} & \underline{0.466} {\scriptsize $\pm$ $3e^{-3}$} & 0.471 {\scriptsize $\pm$ $1e^{-3}$} & 0.478 {\scriptsize $\pm$ $4e^{-3}$} & 0.458 {\scriptsize $\pm$ $3e^{-3}$} & 0.484 {\scriptsize $\pm$ $3e^{-3}$}  \\

            & Avg & \textbf{0.412} {\scriptsize $\pm$ $22e^{-3}$} & \textbf{0.430} {\scriptsize $\pm$ $20e^{-3}$} & \underline{0.426} {\scriptsize $\pm$ $30e^{-3}$} & \underline{0.435} {\scriptsize $\pm$ $25e^{-3}$} & 0.441 {\scriptsize $\pm$ $37e^{-3}$} & 0.439 {\scriptsize $\pm$ $21e^{-3}$} & 0.441 {\scriptsize $\pm$ $29e^{-3}$} & 0.448 {\scriptsize $\pm$ $23e^{-3}$} & \underline{0.426} {\scriptsize $\pm$ $27e^{-3}$} & 0.445 {\scriptsize $\pm$ $26e^{-3}$}   \\
            \midrule

        \multirow{5}{*}{\begin{turn}{90} 
         ETTh2 
         \end{turn}} 
            & 96  & \textbf{0.293} {\scriptsize $\pm$ $8e^{-3}$} & \textbf{0.346} {\scriptsize $\pm$ $5e^{-3}$} & 0.300 {\scriptsize $\pm$ $10e^{-3}$} & \underline{0.351} {\scriptsize $\pm$ $8e^{-3}$} & 0.297 {\scriptsize $\pm$ $5e^{-3}$} & \underline{0.351} {\scriptsize $\pm$ $8e^{-3}$} & 0.300 {\scriptsize $\pm$ $7e^{-3}$} & 0.354 {\scriptsize $\pm$ $6e^{-3}$} & \underline{0.295} {\scriptsize $\pm$ $10e^{-3}$} & 0.353 {\scriptsize $\pm$ $11e^{-3}$} \\
            
            & 192 & \textbf{0.342} {\scriptsize $\pm$ $7e^{-3}$} & \textbf{0.382} {\scriptsize $\pm$ $10e^{-3}$} & \underline{0.348} {\scriptsize $\pm$ $7e^{-3}$} & \underline{0.387} {\scriptsize $\pm$ $9e^{-3}$} & 0.369 {\scriptsize $\pm$ $8e^{-3}$} & 0.388 {\scriptsize $\pm$ $9e^{-3}$} & 0.370 {\scriptsize $\pm$ $12e^{-3}$} & 0.397 {\scriptsize $\pm$ $10e^{-3}$} & 0.410 {\scriptsize $\pm$ $8e^{-3}$} & 0.421 {\scriptsize $\pm$ $11e^{-3}$} \\
            
            & 336 & \underline{0.369} {\scriptsize $\pm$ $8e^{-3}$} & \underline{0.411} {\scriptsize $\pm$ $6e^{-3}$} & \textbf{0.365} {\scriptsize $\pm$ $10e^{-3}$} & \textbf{0.405} {\scriptsize $\pm$ $11e^{-3}$} & 0.422 {\scriptsize $\pm$ $12e^{-3}$} & 0.426 {\scriptsize $\pm$ $18e^{-3}$} & 0.401 {\scriptsize $\pm$ $13e^{-3}$} & 0.426 {\scriptsize $\pm$ $14e^{-3}$} & 0.423 {\scriptsize $\pm$ $7e^{-3}$} & 0.441 {\scriptsize $\pm$ $8e^{-3}$} \\
            
            & 720 & \textbf{0.407} {\scriptsize $\pm$ $3e^{-3}$} & \textbf{0.438} {\scriptsize $\pm$ $5e^{-3}$} & \textbf{0.407} {\scriptsize $\pm$ $4e^{-3}$} & \underline{0.444} {\scriptsize $\pm$ $4e^{-3}$} & \underline{0.421} {\scriptsize $\pm$ $4e^{-3}$} & 0.450 {\scriptsize $\pm$ $6e^{-3}$} & 0.428 {\scriptsize $\pm$ $7e^{-3}$} & 0.455 {\scriptsize $\pm$ $7e^{-3}$} & 0.455 {\scriptsize $\pm$ $3e^{-3}$} & 0.467 {\scriptsize $\pm$ $4e^{-3}$} \\
            
            & Avg & \textbf{0.353} {\scriptsize $\pm$ $41e^{-3}$} & \textbf{0.394} {\scriptsize $\pm$ $34e^{-3}$} & \underline{0.355} {\scriptsize $\pm$ $38e^{-3}$} & \underline{0.397} {\scriptsize $\pm$ $33e^{-3}$} & 0.376 {\scriptsize $\pm$ $51e^{-3}$} & 0.404 {\scriptsize $\pm$ $38e^{-3}$} & 0.374 {\scriptsize $\pm$ $48e^{-3}$} & 0.408 {\scriptsize $\pm$ $37e^{-3}$} & 0.396 {\scriptsize $\pm$ $60e^{-3}$} & 0.420 {\scriptsize $\pm$ $42e^{-3}$} \\
            \midrule

        \multirow{5}{*}{\begin{turn}{90} 
         ETTm1 
         \end{turn}} 
            & 96  & \textbf{0.296} {\scriptsize $\pm$ $4e^{-3}$}  & \textbf{0.350} {\scriptsize $\pm$ $6e^{-3}$} & \underline{0.301} {\scriptsize $\pm$ $8e^{-3}$} & \underline{0.351} {\scriptsize $\pm$ $11e^{-3}$} & 0.324 {\scriptsize $\pm$ $8e^{-3}$} & 0.353 {\scriptsize $\pm$ $6e^{-3}$} & 0.300 {\scriptsize $\pm$ $5e^{-3}$} & 0.353 {\scriptsize $\pm$ $4e^{-3}$} & 0.304 {\scriptsize $\pm$ $4e^{-3}$} & 0.361 {\scriptsize $\pm$ $4e^{-3}$} \\
            
            & 192 & \textbf{0.332} {\scriptsize $\pm$ $7e^{-3}$} & \textbf{0.369} {\scriptsize $\pm$ $10e^{-3}$} & \textbf{0.332} {\scriptsize $\pm$ $9e^{-3}$} & \underline{0.371} {\scriptsize $\pm$ $11e^{-3}$} & 0.369 {\scriptsize $\pm$ $8e^{-3}$} & 0.376 {\scriptsize $\pm$ $9e^{-3}$} & \underline{0.333} {\scriptsize $\pm$ $7e^{-3}$} & 0.377 {\scriptsize $\pm$ $10e^{-3}$} & 0.358 {\scriptsize $\pm$ $5e^{-3}$} & 0.384 {\scriptsize $\pm$ $5e^{-3}$} \\
            
            & 336 & \textbf{0.360} {\scriptsize $\pm$ $4e^{-3}$} & \textbf{0.385} {\scriptsize $\pm$ $5e^{-3}$} & 0.365 {\scriptsize $\pm$ $4e^{-3}$} & \underline{0.391} {\scriptsize $\pm$ $4e^{-3}$} & 0.407 {\scriptsize $\pm$ $4e^{-3}$} & 0.399 {\scriptsize $\pm$ $8e^{-3}$} & \underline{0.362} {\scriptsize $\pm$ $7e^{-3}$} & 0.396 {\scriptsize $\pm$ $6e^{-3}$} & 0.364 {\scriptsize $\pm$ $6e^{-3}$} & 0.397 {\scriptsize $\pm$ $3e^{-3}$} \\
            
            & 720 & \underline{0.411} {\scriptsize $\pm$ $6e^{-3}$} & \textbf{0.418} {\scriptsize $\pm$ $5e^{-3}$} & 0.421 {\scriptsize $\pm$ $7e^{-3}$} & \underline{0.425} {\scriptsize $\pm$ $4e^{-3}$} & 0.472 {\scriptsize $\pm$ $6e^{-3}$} & 0.436 {\scriptsize $\pm$ $11e^{-3}$} & \textbf{0.407} {\scriptsize $\pm$ $8e^{-3}$} & 0.429 {\scriptsize $\pm$ $7e^{-3}$} & 0.424 {\scriptsize $\pm$ $5e^{-3}$} & 0.431 {\scriptsize $\pm$ $4e^{-3}$} \\
            
            & Avg & \underline{0.350} {\scriptsize $\pm$ $42e^{-3}$}& \textbf{0.380} {\scriptsize $\pm$ $25e^{-3}$} & 0.354 {\scriptsize $\pm$ $44e^{-3}$} & \underline{0.385} {\scriptsize $\pm$ $54e^{-3}$} & 0.393 {\scriptsize $\pm$ $31e^{-3}$} & 0.391 {\scriptsize $\pm$ $31e^{-3}$} & \textbf{0.349} {\scriptsize $\pm$ $39e^{-3}$} & 0.389 {\scriptsize $\pm$ $28e^{-3}$} & 0.363 {\scriptsize $\pm$ $43e^{-3}$}  & 0.389 {\scriptsize $\pm$ $25e^{-3}$} \\
            \midrule
            
        \multirow{5}{*}{\begin{turn}{90} 
         ETTm2 
         \end{turn}} 
            & 96  & \underline{0.174} {\scriptsize $\pm$ $3e^{-3}$} & 0.264 {\scriptsize $\pm$ $6e^{-3}$} & \textbf{0.169} {\scriptsize $\pm$ $7e^{-3}$} & \underline{0.260} {\scriptsize $\pm$ $5e^{-3}$} & 0.178 {\scriptsize $\pm$ $2e^{-3}$} & \textbf{0.255} {\scriptsize $\pm$ $3e^{-3}$} & 0.181 {\scriptsize $\pm$ $7e^{-3}$} & 0.269 {\scriptsize $\pm$ $5e^{-3}$} & 0.178 {\scriptsize $\pm$ $6e^{-3}$} & 0.269 {\scriptsize $\pm$ $6e^{-3}$}  \\
            
            & 192 & \textbf{0.233} {\scriptsize $\pm$ $10e^{-3}$} & \underline{0.300} {\scriptsize $\pm$ $6e^{-3}$} & \underline{0.237} {\scriptsize $\pm$ $8e^{-3}$} & 0.306 {\scriptsize $\pm$ $8e^{-3}$} & 0.243 {\scriptsize $\pm$ $6e^{-3}$} & \textbf{0.298} {\scriptsize $\pm$ $7e^{-3}$} & 0.257 {\scriptsize $\pm$ $7e^{-3}$} & 0.323 {\scriptsize $\pm$ $8e^{-3}$} & 0.241 {\scriptsize $\pm$ $6e^{-3}$} & 0.313 {\scriptsize $\pm$ $8e^{-3}$}  \\
            
            & 336 & \underline{0.280} {\scriptsize $\pm$ $6e^{-3}$} & 0.338 {\scriptsize $\pm$ $2e^{-3}$} & \textbf{0.276} {\scriptsize $\pm$ $13e^{-3}$}  & \textbf{0.328} {\scriptsize $\pm$ $11e^{-3}$} & 0.303 {\scriptsize $\pm$ $16e^{-3}$} & 0.338 {\scriptsize $\pm$ $8e^{-3}$} & 0.293 {\scriptsize $\pm$ $8e^{-3}$} & 0.346 {\scriptsize $\pm$ $9e^{-3}$} & 0.287 {\scriptsize $\pm$ $9e^{-3}$} & \underline{0.337} {\scriptsize $\pm$ $8e^{-3}$}  \\
            
            & 720 & \textbf{0.358} {\scriptsize $\pm$ $8e^{-3}$} & \textbf{0.381} {\scriptsize $\pm$ $4e^{-3}$} & 0.366 {\scriptsize $\pm$ $5e^{-3}$} & \underline{0.390} {\scriptsize $\pm$ $4e^{-3}$} & 0.403 {\scriptsize $\pm$ $4e^{-3}$} & 0.396 {\scriptsize $\pm$ $5e^{-3}$} & 0.367 {\scriptsize $\pm$ $3e^{-3}$} & 0.396 {\scriptsize $\pm$ $5e^{-3}$} & \underline{0.364} {\scriptsize $\pm$ $3e^{-3}$} & 0.401 {\scriptsize $\pm$ $3e^{-3}$}  \\
            
            & Avg & \textbf{0.261} {\scriptsize $\pm$ $67e^{-3}$} & 0.323 {\scriptsize $\pm$ $44e^{-3}$} & \underline{0.262} {\scriptsize $\pm$ $71e^{-3}$} & \textbf{0.321} {\scriptsize $\pm$ $47e^{-3}$} & 0.282 {\scriptsize $\pm$ $83e^{-3}$} & \underline{0.322} {\scriptsize $\pm$ $52e^{-3}$} & 0.279 {\scriptsize $\pm$ $67e^{-3}$} & 0.331 {\scriptsize $\pm$ $46e^{-3}$} & 0.267 {\scriptsize $\pm$ $68e^{-3}$} & 0.330 {\scriptsize $\pm$ $48e^{-3}$} \\
            \midrule

        \multirow{5}{*}{\begin{turn}{90} 
         Illness 
         \end{turn}} 
            & 24   & \underline{1.942} {\scriptsize $\pm$ $46e^{-3}$} & \textbf{0.835} {\scriptsize $\pm$ $30e^{-3}$} & 2.105 {\scriptsize $\pm$ $45e^{-3}$} & 0.908 {\scriptsize $\pm$ $70e^{-3}$} & 2.271 {\scriptsize $\pm$ $81e^{-3}$} & 1.008 {\scriptsize $\pm$ $73e^{-3}$} & 2.622 {\scriptsize $\pm$ $64e^{-3}$} & 1.113 {\scriptsize $\pm$ $87e^{-3}$} & \textbf{1.921} {\scriptsize $\pm$ $48e^{-3}$} & \underline{0.902} {\scriptsize $\pm$ $69e^{-3}$}  \\
            
            & 36   & \textbf{1.952} {\scriptsize $\pm$ $48e^{-3}$} & 0.948 {\scriptsize $\pm$ $66e^{-3}$} & 2.066 {\scriptsize $\pm$ $68e^{-3}$} & 0.962 {\scriptsize $\pm$ $55e^{-3}$} & 2.365 {\scriptsize $\pm$ $58e^{-3}$} & 1.029 {\scriptsize $\pm$ $34e^{-3}$} & 2.515 {\scriptsize $\pm$ $62e^{-3}$} & 1.098 {\scriptsize $\pm$ $46e^{-3}$} & \underline{1.992} {\scriptsize $\pm$ $57e^{-3}$} & 0.918 {\scriptsize $\pm$ $48e^{-3}$} \\
            
            & 48   & \textbf{1.738} {\scriptsize $\pm$ $83e^{-3}$} & \textbf{0.886} {\scriptsize $\pm$ $71e^{-3}$} & \underline{1.902} {\scriptsize $\pm$ $90e^{-3}$} & \underline{0.906} {\scriptsize $\pm$ $97e^{-3}$} & 2.395 {\scriptsize $\pm$ $24e^{-3}$} & 1.046 {\scriptsize $\pm$ $39e^{-3}$} & 2.530 {\scriptsize $\pm$ $22e^{-3}$} & 1.066 {\scriptsize $\pm$ $37e^{-3}$} & 2.187 {\scriptsize $\pm$ $31e^{-3}$} & 1.031 {\scriptsize $\pm$ $12e^{-3}$} \\
            
            & 60   & \textbf{1.907} {\scriptsize $\pm$ $34e^{-3}$} & \textbf{0.942} {\scriptsize $\pm$ $38e^{-3}$} & 2.341 {\scriptsize $\pm$ $46e^{-3}$} & 1.016 {\scriptsize $\pm$ $27e^{-3}$} & 2.415 {\scriptsize $\pm$ $36e^{-3}$} & 1.056 {\scriptsize $\pm$ $20e^{-3}$} & 2.329 {\scriptsize $\pm$ $21e^{-3}$} & 1.065 {\scriptsize $\pm$ $14e^{-3}$} & 2.125 {\scriptsize $\pm$ $27e^{-3}$} & \underline{0.992} {\scriptsize $\pm$ $21e^{-3}$} \\
            
            & Avg  & \textbf{1.885} {\scriptsize $\pm$ $86e^{-3}$} & \textbf{0.902} {\scriptsize $\pm$ $46e^{-3}$} & 2.103 {\scriptsize $\pm$ $157e^{-3}$} & 0.948 {\scriptsize $\pm$ $45e^{-3}$} & 2.361 {\scriptsize $\pm$ $55e^{-3}$} & 1.035 {\scriptsize $\pm$ $18e^{-3}$} & 2.387 {\scriptsize $\pm$ $106e^{-3}$} & 1.052 {\scriptsize $\pm$ $21e^{-3}$} & \underline{2.056} {\scriptsize $\pm$ $102e^{-3}$} & 0.961 {\scriptsize $\pm$ $53e^{-3}$} \\
            \midrule

        \multirow{5}{*}{\begin{turn}{90} 
        Weather 
         \end{turn}} 
            & 96  & \textbf{0.151} {\scriptsize $\pm$ $7e^{-3}$} & \underline{0.200} {\scriptsize $\pm$ $3e^{-3}$} & \underline{0.152} {\scriptsize $\pm$ $2e^{-3}$} & 0.201 {\scriptsize $\pm$ $4e^{-3}$} & 0.163 {\scriptsize $\pm$ $6e^{-3}$} & 0.203 {\scriptsize $\pm$ $9e^{-3}$} & 0.157 {\scriptsize $\pm$ $8e^{-3}$} & 0.209 {\scriptsize $\pm$ $8e^{-3}$} & 0.154 {\scriptsize $\pm$ $7e^{-3}$} & \textbf{0.199} {\scriptsize $\pm$ $6e^{-3}$}  \\
            
            & 192 & \textbf{0.197} {\scriptsize $\pm$ $7e^{-3}$} & \textbf{0.243} {\scriptsize $\pm$ $5e^{-3}$} & \textbf{0.197} {\scriptsize $\pm$ $5e^{-3}$} & \underline{0.244} {\scriptsize $\pm$ $4e^{-3}$} & 0.210 {\scriptsize $\pm$ $4e^{-3}$} & 0.248 {\scriptsize $\pm$ $4e^{-3}$} & \underline{0.200} {\scriptsize $\pm$ $11e^{-3}$} & 0.251 {\scriptsize $\pm$ $9e^{-3}$} & 0.214 {\scriptsize $\pm$ $7e^{-3}$} & 0.254 {\scriptsize $\pm$ $8e^{-3}$}  \\
            
            & 336 & \textbf{0.248} {\scriptsize $\pm$ $10e^{-3}$} & \textbf{0.284} {\scriptsize $\pm$ $13e^{-3}$} & \underline{0.253} {\scriptsize $\pm$ $7e^{-3}$} & \underline{0.289} {\scriptsize $\pm$ $5e^{-3}$} & 0.266 {\scriptsize $\pm$ $6e^{-3}$} & 0.290 {\scriptsize $\pm$ $5e^{-3}$} & \underline{0.253} {\scriptsize $\pm$ $7e^{-3}$} & 0.288 {\scriptsize $\pm$ $8e^{-3}$} & 0.273 {\scriptsize $\pm$ $6e^{-3}$} & 0.291 {\scriptsize $\pm$ $11e^{-3}$} \\
            
            & 720 & \textbf{0.330} {\scriptsize $\pm$ $3e^{-3}$} & \underline{0.340} {\scriptsize $\pm$ $2e^{-3}$} & \underline{0.332} {\scriptsize $\pm$ $4e^{-3}$} & \underline{0.341} {\scriptsize $\pm$ $7e^{-3}$} & 0.347 {\scriptsize $\pm$ $6e^{-3}$} & 0.344 {\scriptsize $\pm$ $6e^{-3}$} & 0.338 {\scriptsize $\pm$ $2e^{-3}$} & \textbf{0.333} {\scriptsize $\pm$ $4e^{-3}$} & 0.364 {\scriptsize $\pm$ $7e^{-3}$} & 0.350 {\scriptsize $\pm$ $4e^{-3}$} \\
            
            & Avg & \textbf{0.231} {\scriptsize $\pm$ $66e^{-3}$} & \textbf{0.266} {\scriptsize $\pm$ $46e^{-3}$} & \underline{0.233} {\scriptsize $\pm$ $68e^{-3}$} & \underline{0.269} {\scriptsize $\pm$ $52e^{-3}$} & 0.246 {\scriptsize $\pm$ $68e^{-3}$} & 0.271 {\scriptsize $\pm$ $52e^{-3}$} & 0.237 {\scriptsize $\pm$ $68e^{-3}$} & 0.270 {\scriptsize $\pm$ $29e^{-3}$} & 0.251 {\scriptsize $\pm$ $78e^{-3}$} & 0.274 {\scriptsize $\pm$ $54e^{-3}$} \\
            \midrule

        \multirow{5}{*}{\begin{turn}{90} 
         Traffic 
         \end{turn}} 
            & 96  & \textbf{0.369} {\scriptsize $\pm$ $10e^{-3}$} & 0.266 {\scriptsize $\pm$ $11e^{-3}$} & \underline{0.372} {\scriptsize $\pm$ $6e^{-3}$} & \textbf{0.255} {\scriptsize $\pm$ $9e^{-3}$} & 0.433 {\scriptsize $\pm$ $5e^{-3}$} & 0.292 {\scriptsize $\pm$ $7e^{-3}$} & 0.379 {\scriptsize $\pm$ $2e^{-3}$} & 0.274 {\scriptsize $\pm$ $5e^{-3}$} & 0.388 {\scriptsize $\pm$ $4e^{-3}$} & \underline{0.261} {\scriptsize $\pm$ $7e^{-3}$} \\
            
            & 192 & \textbf{0.391} {\scriptsize $\pm$ $4e^{-3}$} & \underline{0.277} {\scriptsize $\pm$ $3e^{-3}$} & 0.399 {\scriptsize $\pm$ $6e^{-3}$} & 0.284 {\scriptsize $\pm$ $3e^{-3}$} & 0.451 {\scriptsize $\pm$ $8e^{-3}$} & 0.297 {\scriptsize $\pm$ $10e^{-3}$} & \underline{0.397} {\scriptsize $\pm$ $9e^{-3}$} & 0.282 {\scriptsize $\pm$ $11e^{-3}$} & 0.409 {\scriptsize $\pm$ $7e^{-3}$} & \textbf{0.272} {\scriptsize $\pm$ $7e^{-3}$}  \\
            
            & 336 & \textbf{0.408} {\scriptsize $\pm$ $4e^{-3}$} & \underline{0.287} {\scriptsize $\pm$ $3e^{-3}$} & 0.413 {\scriptsize $\pm$ $5e^{-3}$} & \textbf{0.273} {\scriptsize $\pm$ $7e^{-3}$} & 0.467 {\scriptsize $\pm$ $10e^{-3}$} & 0.302 {\scriptsize $\pm$ $11e^{-3}$} & \underline{0.410} {\scriptsize $\pm$ $7e^{-3}$} & 0.289 {\scriptsize $\pm$ $4e^{-3}$} & 0.438 {\scriptsize $\pm$ $5e^{-3}$} & 0.299 {\scriptsize $\pm$ $3e^{-3}$} \\
            
            & 720 & \textbf{0.435} {\scriptsize $\pm$ $3e^{-3}$} & \underline{0.298} {\scriptsize $\pm$ $3e^{-3}$} & \underline{0.440} {\scriptsize $\pm$ $4e^{-3}$} & \textbf{0.291} {\scriptsize $\pm$ $2e^{-3}$} & 0.485 {\scriptsize $\pm$ $7e^{-3}$} & 0.323 {\scriptsize $\pm$ $5e^{-3}$} & 0.445 {\scriptsize $\pm$ $4e^{-3}$} & 0.301 {\scriptsize $\pm$ $8e^{-3}$} & 0.462 {\scriptsize $\pm$ $6e^{-3}$} & 0.301 {\scriptsize $\pm$ $4e^{-3}$} \\
            
            & Avg & \textbf{0.400} {\scriptsize $\pm$ $17e^{-3}$} & \underline{0.282} {\scriptsize $\pm$ $12e^{-3}$} & \underline{0.406} {\scriptsize $\pm$ $25e^{-3}$} & \textbf{0.276} {\scriptsize $\pm$ $16e^{-3}$} & 0.459 {\scriptsize $\pm$ $19e^{-3}$} & 0.303 {\scriptsize $\pm$ $12e^{-3}$} & 0.407 {\scriptsize $\pm$ $24e^{-3}$} & 0.287 {\scriptsize $\pm$ $10e^{-3}$} & 0.424 {\scriptsize $\pm$ $28e^{-3}$} & 0.283 {\scriptsize $\pm$ $17e^{-3}$} \\
            \midrule
            
        \multirow{5}{*}{\begin{turn}{90} 
         ECL 
         \end{turn}} 
            & 96  & \textbf{0.128} {\scriptsize $\pm$ $8e^{-3}$} & \textbf{0.221} {\scriptsize $\pm$ $5e^{-3}$} & 0.133 {\scriptsize $\pm$ $4e^{-3}$} & 0.228 {\scriptsize $\pm$ $6e^{-3}$} & 0.167 {\scriptsize $\pm$ $7e^{-3}$} & 0.254 {\scriptsize $\pm$ $3e^{-3}$} & \underline{0.131} {\scriptsize $\pm$ $4e^{-3}$} & \underline{0.226} {\scriptsize $\pm$ $10e^{-3}$} & 0.172 {\scriptsize $\pm$ $8e^{-3}$} & 0.265 {\scriptsize $\pm$ $7e^{-3}$} \\
            
            & 192 & \textbf{0.151} {\scriptsize $\pm$ $9e^{-3}$} & \textbf{0.247} {\scriptsize $\pm$ $5e^{-3}$} & \underline{0.157} {\scriptsize $\pm$ $4e^{-3}$} & \underline{0.250} {\scriptsize $\pm$ $3e^{-3}$} & 0.178 {\scriptsize $\pm$ $5e^{-3}$} & 0.264 {\scriptsize $\pm$ $4e^{-3}$} & 0.158 {\scriptsize $\pm$ $3e^{-3}$} & 0.252 {\scriptsize $\pm$ $3e^{-3}$} & 0.182 {\scriptsize $\pm$ $6e^{-3}$} & 0.279 {\scriptsize $\pm$ $5e^{-3}$}  \\
            
            & 336 & \textbf{0.170} {\scriptsize $\pm$ $7e^{-3}$} & \textbf{0.269} {\scriptsize $\pm$ $4e^{-3}$} & \underline{0.176} {\scriptsize $\pm$ $10e^{-3}$} & 0.281 {\scriptsize $\pm$ $13e^{-3}$} & 0.196 {\scriptsize $\pm$ $5e^{-3}$} & 0.282 {\scriptsize $\pm$ $7e^{-3}$} & 0.183 {\scriptsize $\pm$ $6e^{-3}$} & \underline{0.280} {\scriptsize $\pm$ $3e^{-3}$} & 0.195 {\scriptsize $\pm$ $2e^{-3}$} & 0.288 {\scriptsize $\pm$ $4e^{-3}$}  \\
            
            & 720 & \textbf{0.207} {\scriptsize $\pm$ $4e^{-3}$} & \textbf{0.299} {\scriptsize $\pm$ $3e^{-3}$} & 0.215 {\scriptsize $\pm$ $5e^{-3}$} & \underline{0.302} {\scriptsize $\pm$ $9e^{-3}$} & 0.232 {\scriptsize $\pm$ $4e^{-3}$} & 0.312 {\scriptsize $\pm$ $4e^{-3}$} & \underline{0.210} {\scriptsize $\pm$ $4e^{-3}$} & 0.306 {\scriptsize $\pm$ $3e^{-3}$} & 0.233 {\scriptsize $\pm$ $7e^{-3}$} & 0.320 {\scriptsize $\pm$ $5e^{-3}$} \\
            
            & Avg & \textbf{0.164} {\scriptsize $\pm$ $29e^{-3}$} & \textbf{0.256} {\scriptsize $\pm$ $29e^{-3}$} & \underline{0.170} {\scriptsize $\pm$ $30e^{-3}$} & \underline{0.265} {\scriptsize $\pm$ $28e^{-3}$} & 0.193 {\scriptsize $\pm$ $25e^{-3}$} & 0.278 {\scriptsize $\pm$ $22e^{-3}$} & \underline{0.170} {\scriptsize $\pm$ $29e^{-3}$} & 0.266 {\scriptsize $\pm$ $30e^{-3}$} & 0.195 {\scriptsize $\pm$ $23e^{-3}$} & 0.288 {\scriptsize $\pm$ $20e^{-3}$} \\
            \midrule
         \multicolumn{2}{c|}{Wins} & \multicolumn{2}{c|}{ 51 } & \multicolumn{2}{c|}{ 14 } & \multicolumn{2}{c|}{ 1 } & \multicolumn{2}{c|}{ 3 } & \multicolumn{2}{c|}{ 3 } \\
         \bottomrule
    \end{tabular}
    }
    \caption{Full Results Long-term Forecasting with LLM4TS with std: \textbf{Bold Text} indicate the best performer and \underline{Underline Text} indicate the second best performer. We count the number of wins by the number of best performer.}
    \label{tab:LLM4TS_with_std}
\end{table*}

\end{document}